\documentclass[review]{elsarticle}

\usepackage{amsthm}
\usepackage{lineno,hyperref}

\usepackage{amssymb}
\usepackage{color}
\usepackage[fleqn]{amsmath}
\usepackage{graphicx}
\usepackage{float}
\usepackage{multirow}
\usepackage{caption}
\usepackage{array}
\usepackage{rotating}
\usepackage{arydshln}
\usepackage{makeidx}
\usepackage{subfigure}

\usepackage{multicol}

\usepackage{stfloats}

\usepackage{subeqnarray}
\usepackage{cases}

\makeindex

\newcommand{\MYnextline}[2]{\begin{tabular}{@{}#1@{}}#2\end{tabular}}

\newcommand {\SCB}[0]{0.95}

\makeatletter

\theoremstyle{plain}
\newtheorem{thm}{Theorem}[section]

\theoremstyle{remark}

\journal{\textbf{Pattern Recognition}}

\begin{document}

\begin{frontmatter}

\title{\textbf{Joint Gender Classification and Age Estimation by Nearly Orthogonalizing Their Semantic Spaces}}

\author[nuaa]{Qing Tian}

\author[nuaa]{Songcan Chen\corref{myCorrespondingAuthor}}
\cortext[myCorrespondingAuthor]{Corresponding author}
\ead{s.chen@nuaa.edu.cn}

\address[nuaa]{College of Computer Science and Technology, Nanjing University of Aeronautics and Astronautics, Nanjing 210016, China}

\begin{abstract}
In human face-based biometrics, gender classification and age estimation are two typical learning tasks. Although a variety of approaches have been proposed to handle them, just a few of them are solved jointly, even so, these joint methods do not yet specifically concern the semantic difference between human gender and age, which is intuitively helpful for joint learning, consequently leaving us a room of further improving the performance. To this end, in this work we firstly propose a general learning framework for jointly estimating human gender and age by specially attempting to formulate such semantic relationships as a form of near-orthogonality regularization and then incorporate it into the objective of the joint learning framework. In order to evaluate the effectiveness of the proposed framework, we exemplify it by respectively taking the widely used binary-class SVM for gender classification, and two threshold-based ordinal regression methods (i.e., the discriminant learning for ordinal regression and support vector ordinal regression) for age estimation, and crucially coupling both through the proposed semantic formulation. Moreover, we develop its kernelized nonlinear counterpart by deriving a representer theorem for the joint learning strategy. Finally, through extensive experiments on three aging datasets FG-NET, Morph Album I and Morph Album II, we demonstrate the effectiveness and superiority of the proposed joint learning strategy.
\end{abstract}


\end{frontmatter}


\section{Introduction} \label{sec:introduction}
Human face conveys various biometric traits, such as gender, age, ethnicity, and expression, in which the estimations of face-based gender and/or age have attracted increasing attentions due to their wide real-world applicability in recommendation systems \cite{fjermestad2006electronic}, \cite{linoff2011data}, \cite{raab2012customer}, security access control \cite{lanitis2004comparing}, \cite{ramanathan2006face}, \cite{LARR_guo2008image}, \cite{wu2013attribute}, biometrics \cite{van2004dependency}, \cite{patterson2007aspects}, entertainment \cite{das2003automatic}, \cite{gallagher2009understanding}, \cite{chen2013travel}, \cite{dibeklioglu2015recognition}, and cosmetology \cite{yun2004facetransfer}, \cite{fu2006m}, etc. Therefore, in this work, we concentrate on the problem of joint estimation for human gender and age. Note that the proposed joint learning methods can act as a pioneering reference to develop methodologies for other joint estimations, e.g., age/expression joint estimation.

\textbf{Gender Classification} is usually handled as a typical binary classification problem and implemented using off-the-shelf binary classifiers, such as the neural networks based methods \cite{golomb1990sexnet}, \cite{poggio1992hyberbf}, \cite{phillips2000feret}, the SVM-based methods \cite{moghaddam2002learning}, \cite{graf2002gender}, \cite{jain2004integrating}, \cite{costen2004sparse}, \cite{makinen2008evaluation}, and the Boosting-related approaches \cite{shakhnarovich2002unified}, \cite{wu2003lut}, \cite{baluja2007boosting}, \cite{xu2007soda}, Bayesian Classifier \cite{shih2013robust}, Random Forest based method \cite{xia2015combining}, etc. Besides the above representative methods, there have been other types of methods proposed for gender classification (see \cite{wu2015human}).

\textbf{Age Estimation} is relatively more challenging to evaluate, compared to the binary gender recognition, due to its multi-class characteristic, especially the ordinality in value. As a result, researchers have paid more attention on it and proposed a variety of methods. These methods can be roughly categorized into three families: classification-based (e.g., \cite{lanitis2004comparing}, \cite{geng2013facial}, \cite{ueki2006subspace}, \cite{alnajar2012learning}, \cite{sai2015facial}, \cite{alnajar2014expression}, \cite{dibeklioglu2015combining}), regression-based (e.g., \cite{lanitis2002toward}, \cite{fu2007estimating}, \cite{luu2009age}, \cite{yan2007ranking}, \cite{yan2007learning}, \cite{geng2007automatic}, \cite{fu2008human}, \cite{chang2011ordinal}, \cite{ni2011web}, \cite{li2012learning_ICPRmetric}, \cite{li2012learning_learnforfeature}, \cite{chao2013facial}), and their hybrid methods (e.g., \cite{LARR_guo2008image}, \cite{kohli2013hierarchical}). When we treat each age as a generic class and thus we can perform age estimation using the multi-class classification framework. Along this line, the artificial neural networks (ANN) \cite{lanitis2004comparing}, conditional probability neural networks (CPNN) \cite{geng2013facial}, Gaussian mixture models \cite{ueki2006subspace}, and extreme learning machines (ELM) \cite{sai2015facial} have been successively employed to age estimation. More recently, \cite{alnajar2014expression} proposed to perform age classification by eliminating the variance of expressions. And, \cite{dibeklioglu2015combining} incorporated facial dynamics together with the facial appearance information for age estimation. Actually, age estimation is more of a regression rather than classification problem due to its natural characteristics of continuity and monotonicity in aging. Along this line, the quadratic function \cite{lanitis2002toward}, \cite{fu2008human}, multiple linear regression \cite{fu2007estimating}, $\xi$-SVR \cite{luu2009age}, SDP regressors \cite{yan2007ranking, yan2007learning}, aging pattern subspace (AGES) \cite{geng2007automatic}, multi-instance regressor \cite{ni2011web}, and KNN-SVR \cite{chao2013facial} have been proposed to perform human age regression. Besides the separate classification or regression based methods, \cite{LARR_guo2008image} established a so-called locally adjusted robust regression (LARR) by assembling a series of classifiers and regressors, and obtained more competitive age estimation results.

\textbf{Joint Estimation of Human Gender and Age} is to estimate human gender and age jointly, apart from aforementioned separate (or say single) gender classification and age estimation. In real world applications, there is usually a requirement of joint prediction for human gender and age, where the gender information as well as the age of a potential customer is desired in such applications as commodity recommendation \cite{fjermestad2006electronic}, \cite{linoff2011data}, \cite{raab2012customer}. To this end, \cite{guo2013joint} and \cite{guo2014framework} employed the off-the-shelf Partial Least Squares (PLS), a kind of multi-output regressor, to make joint estimation for gender and age, as well as the ethnicity. And on the Morph II aging database, they obtained preferable estimation results. However, the PLS is just a generic off-the-shelf regressor, which has not definitely considered the underlying relationship between the gender and age, as claimed in \cite{guo2009study}, \cite{guo2009gender}, and what is worse is that the semantic discrepancy between the gender and age is not reflected between the responses of the PLS. In addition, to perform feature selection for age estimation, \cite{liang2011multi} adopted the multi-task learning framework by taking male-oriented and female-oriented age estimations as two tasks and enforcing group-lasso for group feature selection, provided that the gender labels of training data are known. Although on aging datasets like FG-NET they obtained competitive results based on the selected feature subset, it is essentially a pseudo-joint method, i.e., the gender information is just used as auxiliary for age estimation, since that it makes estimation just for human age rather than the gender and age together, leading to the ignorance of their mutual relationship, and that its two-step strategy (i.e., \emph{multi-task based feature selection} and then \emph{ridge regression based age estimation}) is practically complex and time-consuming. Besides, \cite{yi2015age} adopted the highly-nonlinear convolutional networks to optimize the loss functions of gender, age and ethnicity jointly with sharing the same deep convolutional network structures, and on the Morph II dataset obtained more competitive estimation results. However, the work of \cite{yi2015age} just shares a common deep network learning structure without considering the inherent interactive semantic relationship between the gender and age, since in which the loss functions with respect to the gender and age are just simply accumulated together for optimization without coupling terms.

\vspace{10pt}
\renewcommand\arraystretch{1.2}
\begin{table}[thbp!]
\centering
\caption{Comparison between existing methods and ours in terms of joint estimation for human gender and age.}\label{tab:drawbacks-of-existing-methods}
  \scalebox{0.44}
  {
\begin{tabular}{ccccccc}
\hline
\multicolumn{1}{|c||}{ } & \multicolumn{2}{c|}{Single methods} & \multicolumn{4}{c|}{Joint methods} \\
\hline
\hline
\multicolumn{1}{|c||}{\textbf{Item}} & \multicolumn{1}{c}{\MYnextline{c}{Separate gender estimation \\(\cite{golomb1990sexnet}, \cite{poggio1992hyberbf}, \cite{phillips2000feret}, \cite{moghaddam2002learning}, \cite{graf2002gender}, \cite{jain2004integrating}, \cite{costen2004sparse},\\\cite{makinen2008evaluation}, \cite{shakhnarovich2002unified}, \cite{wu2003lut}, \cite{baluja2007boosting}, \cite{shih2013robust}, \cite{xu2007soda}, \cite{xia2015combining}.)}} & \multicolumn{1}{|c|}{\MYnextline{c}{Separate age estimation\\(\cite{lanitis2004comparing}, \cite{geng2013facial}, \cite{ueki2006subspace}, \cite{alnajar2012learning}, \cite{sai2015facial}, \cite{alnajar2014expression}, \cite{dibeklioglu2015combining}, \\ \cite{lanitis2002toward}, \cite{fu2007estimating}, \cite{luu2009age}, \cite{yan2007ranking}, \cite{yan2007learning}, \cite{geng2007automatic}, \cite{fu2008human},\\ \cite{chang2011ordinal}, \cite{ni2011web}, \cite{li2012learning_ICPRmetric}, \cite{li2012learning_learnforfeature}, \cite{chao2013facial}, \cite{LARR_guo2008image}, \cite{kohli2013hierarchical})}} & \multicolumn{1}{c|}{\MYnextline{c}{PLS-based\\(\cite{guo2013joint}, \cite{guo2014framework})}} & \multicolumn{1}{c|}{\MYnextline{c}{Multi-task based\\feature selection\\(\cite{liang2011multi})}} & \multicolumn{1}{c|}{\MYnextline{c}{Multi-task based\\deep networks\\(\cite{yi2015age})}}& \multicolumn{1}{c|}{\MYnextline{c}{\textbf{\emph{Ours}}}}\\
\hline
\multicolumn{1}{|c||}{\MYnextline{c}{Estimating the gender and age jointly?}} & \multicolumn{1}{c}{\emph{$\times$}} & \multicolumn{1}{|c|}{\emph{$\times$}} & \multicolumn{1}{c|}{\emph{$\surd$}} & \multicolumn{1}{c|}{\emph{$\times$}} & \multicolumn{1}{c|}{\emph{$\surd$}} & \multicolumn{1}{c|}{\emph{$\surd$}}\\
\hline
\multicolumn{1}{|c||}{\MYnextline{c}{Strategically considering\\the underlying correlation\\between the gender and age?}} & \multicolumn{1}{c}{\emph{$\times$}} & \multicolumn{1}{|c|}{\emph{$\times$}} & \multicolumn{1}{c|}{\emph{$\times$}} & \multicolumn{1}{c|}{\emph{$\times$}} & \multicolumn{1}{c|}{\emph{$\times$}} & \multicolumn{1}{c|}{\emph{$\surd$}}\\
\hline
\multicolumn{1}{|c||}{\MYnextline{c}{Considering the ordinal characteristic\\of aging sequence?}} & \multicolumn{1}{c}{\emph{--}} & \multicolumn{1}{|c|}{\emph{$\times$}} & \multicolumn{1}{c|}{\emph{$\times$}} & \multicolumn{1}{c|}{\emph{$\times$}} & \multicolumn{1}{c|}{\emph{$\times$}} & \multicolumn{1}{c|}{\emph{$\surd$}}\\
\hline
\multicolumn{1}{|c||}{\MYnextline{c}{Considering the semantic discrepancy\\between the gender and age?}} & \multicolumn{1}{c}{\emph{$\times$}} & \multicolumn{1}{|c|}{\emph{$\times$}} & \multicolumn{1}{c|}{\emph{$\times$}} & \multicolumn{1}{c|}{\emph{$\times$}} & \multicolumn{1}{c|}{\emph{$\times$}} & \multicolumn{1}{c|}{\emph{$\surd$}}\\
\hline
\end{tabular}
  }
\end{table}

To overcome the drawbacks of the methods aforementioned, in this work we propose a general joint learning framework for human gender and age, in which gender estimation is took as a binary classification and age prediction as an ordinal regression problem. More crucially, as a key ingredient of reflecting the semantic discrepancy between human gender and age, the underlying relationship between their semantic spaces is formulated as a near-orthogonality regularizer and incorporated to regularize the learning. In order to evaluate the effectiveness of the proposed framework, we exemplify it by respectively taking the widely used binary-class SVM for gender classification, and the discriminant learning for ordinal regression/support vector ordinal regression for age estimation, and particularly coupling them with the proposed regularization manner. Then we kernelize the joint learning strategy by deriving a representer theorem and using kernel trick. Finally, through extensive experiments on three real-world aging datasets, we demonstrate the effectiveness and superiority of the proposed joint learning strategy, compared to related methods. And to provide an intuitive illustration about the superiority of the proposed joint learning strategy over existing methods in terms of estimation for human gender and age, we compare them in Table \ref{tab:drawbacks-of-existing-methods}.

The rest of this paper is organized as follows. In Section \ref{sec:related work}, we make a brief review for related works. Then, we propose our methodology for joint estimation of human gender and age in Section \ref{sec:proposed-methodology}. In Section \ref{sec:experiment}, we conduct estimation experiments to evaluate the effectiveness and superiority of the proposed strategy. Finally, we conclude this paper in Section \ref{sec:conclusions}.

\section{Related Work} \label{sec:related work}
Before presenting our methodology, in this section we briefly review some works mostly related to joint estimation for human gender and age.

\subsection{Joint Age Estimation for Males and Females in  Multi-Task Framework} \label{sec:relatedwork-multitask}
In order to conduct joint age estimation for males and females with feature selection, \cite{liang2011multi} adopted the multi-task framework regularized with the group-lasso strategy \cite{meier2008group} by taking male-oriented and female-oriented age estimations as two tasks. More specifically, for a given training set $\{x_{i}^{t}, y_{i}^{t}\} \in \mathcal{R}^{D} \times \mathcal{R}, i=1,...,N_{t}, t=1,2$, where $x_{i}^{t}$ and $y_{i}^{t}$ denote the $i$-th $D$-dimensional instance and its label from the $t$-th task, respectively, they performed joint age estimation for males and females through using the off-the-shelf multi-task feature selection learning which is formulated as
\begin{equation}\label{eq:multi-task}
    \scalebox{\SCB}
    {$
\begin{split}
& \min_{W=[w^{1}, w^{2}]} \; \frac{1}{\sum_{t=1}^{2}N_{t}}\sum_{t=1}^{2}\sum_{i=1}^{N_{t}}\|y_{i}^{t} - (w^{t})^{T} x_{i}^{t}\|_{2}^{2} + \lambda\sum_{d=1}^{D}\|w_{d}\|_{2},
\end{split}
    $}
\end{equation}
where $W=[w^{1}, w^{2}] \in \mathcal{R}^{D\times2}$ consists of the $w^{1}$ and $w^{2}$ which respectively represent the projection weights for the two tasks (corresponding to the male-oriented and female-oriented age estimations), and the second term in \eqref{eq:multi-task} is the well-known group-lasso regularization and is usually used for joint feature selection among the tasks.

Taking the selected features learned by \eqref{eq:multi-task} as new feature representation, Liang et al. uniformly adopted the Ridge Regression (RR) to perform age regression for the males and females, and obtained better results than single-task based methods \cite{geng2007automatic}, \cite{fang2010discriminant}, \cite{guo2009human}.

\subsection{Joint Estimation for Gender and Age using Partial Least Squares (PLS)} \label{sec:relatedwork-pls}
Strictly speaking, one typical work for joint estimation of human gender and age is the one performed by \cite{guo2013joint} and \cite{guo2014framework}, where a popular multi-output regressor, i.e., the PLS, was employed for the joint estimation. And on a large aging dataset, Morph II, they obtained competitive prediction results for gender, age as well as ethnicity.

As for PLS, it is a typical multi-output regressor which can model the mapping from the input features to the responses. More concretely, for given input matrix $X$ and output response matrix $Y$ (whose each response vector is concatenated from the multi-output variables, e.g., the gender and age), both of which are centered with zero-means, the PLS aims to compute two weight vectors, $w$ and $c$, to maximize the following covariance
\begin{equation}\label{eq:pls-cov(Xtw,Ytc)}
    \scalebox{\SCB}
    {$
\begin{split}
& cov(t,u) \triangleq max_{|w|=|c|=1} cov(X^{T}w, Y^{T}c),
\end{split}
    $}
\end{equation}
where $cov(t,u)$ represents the covariance between the score vectors $t\triangleq X^{T}w$ and $u\triangleq Y^{T}c$. Regressions can then be performed for both $X$ and $Y$ based on the score vectors $t$ and $u$ computed by optimizing the \eqref{eq:pls-cov(Xtw,Ytc)}, such that
\begin{equation}\label{eq:pls-XYregressiont1u1}
    \scalebox{\SCB}
    {$
\begin{split}
& X = pt^{T} + X_{1}, \quad Y = qu^{T} + Y_{1},
\end{split}
    $}
\end{equation}
where $p\triangleq\frac{Xt}{t^{T}t}$ and $q\triangleq\frac{Yu}{u^{T}u}$ are called loading vectors, and $X_{1}$ and $Y_{1}$ are the regression residuals of $X$ and $Y$, respectively. Then, based on \eqref{eq:pls-XYregressiont1u1}, we can compute a sequence of score and loading vectors to make the regression residuals small enough as
\begin{equation}\label{eq:pls-XYregressiontNuN}
    \scalebox{\SCB}
    {$
\begin{split}
& X = p_{1}t_{1}^{T} + \cdot\cdot\cdot + p_{k}t_{k}^{T} + X_{k}, \quad Y = q_{1}u_{1}^{T} + \cdot\cdot\cdot + q_{k}u_{k}^{T} + Y_{k}.
\end{split}
    $}
\end{equation}
According to the regression relationships between the loading vectors $T = \{t_{1}, \cdot\cdot\cdot, t_{k}\}$ and $U = \{u_{1}, \cdot\cdot\cdot, u_{k}\}$ (\cite{rosipal2006overview}), we have
\begin{equation}\label{eq:pls-YofX}
    \scalebox{\SCB}
    {$
\begin{split}
& Y = B^{T}X + R_{Y},
\end{split}
    $}
\end{equation}
where $B \triangleq XU(T^{T}X^{T}XU)^{-1}T^{T}Y^{T}$, and $R_{Y}$ stands for the regression residual. Actually, if the number $k$ of iterations is properly assigned, the residual $R_{Y}$ can be omitted for practical applicability.
\\
\\
\textbf{Remark:} It can be found from Section \ref{sec:relatedwork-multitask} that in the multi-task based age estimation, the gender information is just used as auxiliary to help select features for consequent age estimation, rather than estimated jointly with age. Therefore, it is essentially not a real joint method. As for the PLS based joint method reviewed in Section \ref{sec:relatedwork-pls}, when it is directly employed to perform joint estimation for human gender and age as in \cite{guo2013joint} and \cite{guo2014framework}, its two-dimensional output indicates the regressed gender and age results, respectively. However, by this way the heterogenicity between the discrete binaryness of gender and the continuous ordinality of age is seriously destroyed. In addition, as mentioned in Section \ref{sec:introduction} and demonstrated in Table \ref{tab:drawbacks-of-existing-methods}, neither the multi-task based (see Section \ref{sec:relatedwork-multitask}) nor the PLS-based (see Section \ref{sec:relatedwork-pls}) method has respected the ordinal characteristic of human age, or the semantic discrepancy between human gender and age. Consequently, their estimation results are sub-optimal and a more preferable joint estimation strategy for human gender and age is desired.

\section{Proposed Methodology} \label{sec:proposed-methodology}
In this section, we firstly attempt to propose a general learning framework for joint estimation of human gender and age, in which the binaryness of human gender and ordinality of human age are both considered, and in particular, the semantic difference relationship between the human gender and age is exploited and modeled to improve their estimations. Then, we come to exemplify the framework for the following empirical evaluation.

\subsection{A Novel Joint Framework for Human Gender Classification and Age Estimation: Learning in Nearly Orthogonal Semantic Spaces} \label{sec:NOSSpaces}
\begin{figure}[htdp!]
  \centering
  \includegraphics[width=0.7\textwidth]{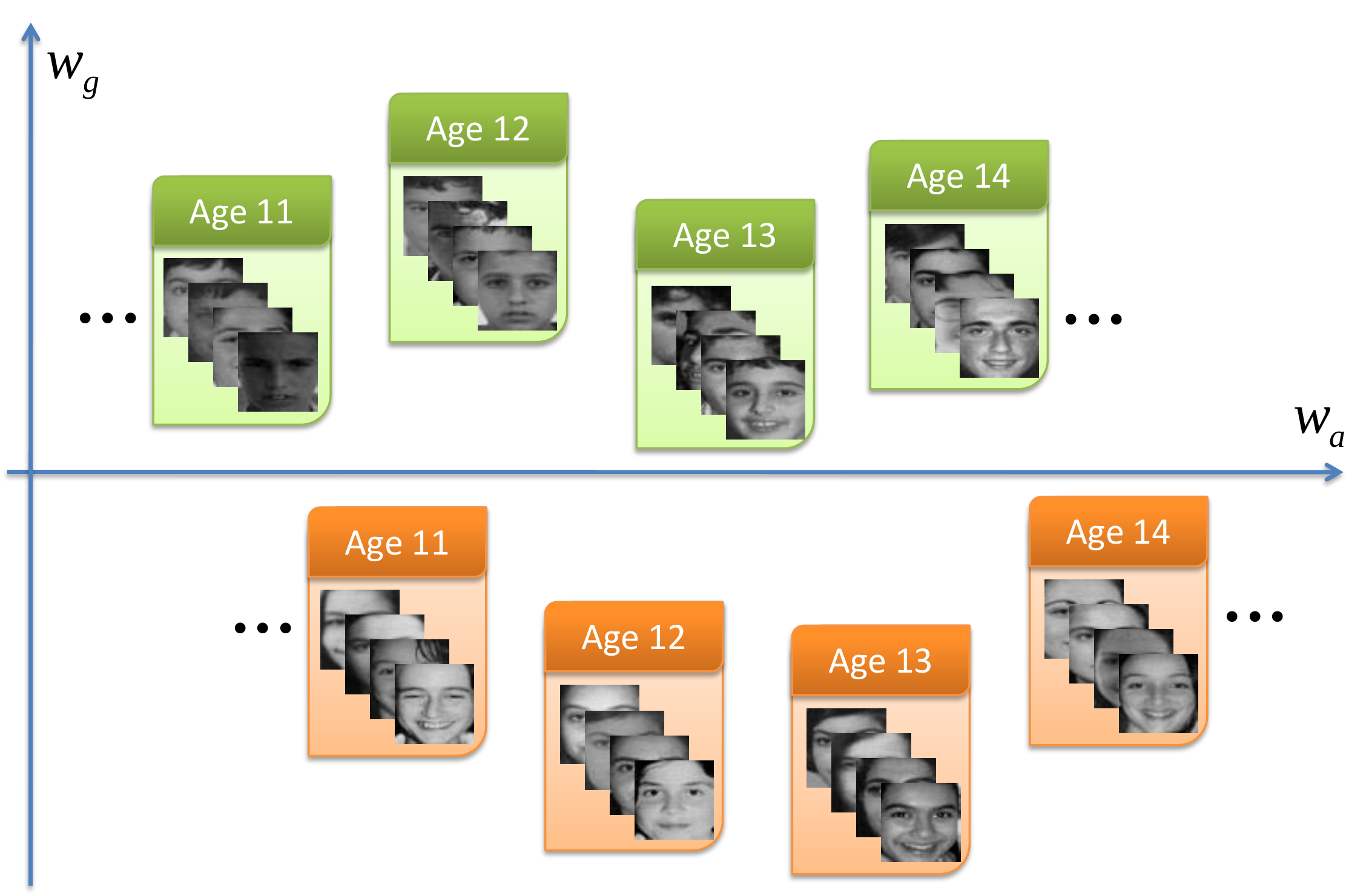}\\
  \caption{Illustration of the proposed joint estimation strategy for human gender and age. The $w_{g}$ indicates the discriminant direction for gender (from females to males), while the $w_{a}$ represents the age regression direction (from younger to elder) \protect\footnotemark.}\label{Fig:illustration_of_joint}
\end{figure}
\footnotetext{Note that the $w_g$ and $w_a$ are not required to be strictly orthogonal to each other, which will be verified in Section \ref{sec:angle analysis}.}

Now consider the situation that there is a set of human face samples associated with gender labels and age annotations, what we need do is to train a joint estimator in terms of the gender and age using the training set. For the sake of clarification, as illustrated in Figure \ref{Fig:illustration_of_joint}, we let $w_{a}$ and $w_{g}$ denote the gender discriminant and age regression directions, respectively. In order to make the samples separated respectively on the negative and positive sub-axes along the $w_{g}$ in terms of gender label, while in positive (for male) or negative (for female) sub-axis of $w_{g}$ the instances distributed ordinal along the $w_{a}$ with respect to the age value, the $w_{g}$ and $w_{a}$ should be generally located as orthogonal as possible to each other, as illustrated in Figure \ref{Fig:illustration_of_joint}. By this way, gender classification along the $w_{g}$ and age estimation along the $w_{a}$ can be performed nearly without mutual collision, and more importantly, the semantic discrepancy between the gender and the age can be eliminated from each other in their individual discriminant or regression direction. That is, the gender can be classified along the $w_{g}$ without the need of considering the age variations, and vice versa. As a result, the estimation accuracy of gender classification and age regression can be improved. For simplicity of constructing a joint learning framework in terms of human gender and age, we mathematically formulate learning in such \emph{nearly orthogonal semantic spaces} (NOSSpaces) as a regularization term
\begin{equation}\label{eq:R-NOSSpaces}
    \scalebox{\SCB}
    {$
\begin{split}
& \mathcal{R}_{NOSSpaces}:= (w_{a}^{T}w_{g})^{2} \\
& \;\quad \qquad \qquad = w_{g}^{T}w_{a}w_{a}^{T}w_{g} \\
& \;\quad \qquad \qquad = w_{a}^{T}w_{g}w_{g}^{T}w_{a}
\end{split}
    $}
\end{equation}
Clearly, according to the above analysis the $\mathcal{R}_{NOSSpaces}$ should be as small as possible to achieve the desired joint estimation for the gender and age, as illustrated in Figure \ref{Fig:illustration_of_joint}.

With the proposed joint learning regularization term $\mathcal{R}_{NOSSpaces}$, we come to propose a general joint learning framework for human gender classification and age estimation as
\begin{equation}\label{eq:joint-framework}
    \scalebox{\SCB}
    {$
\begin{split}
& min_{\{w_{g}, w_{a}\}} \\
& \quad \mathcal{L}_{g}(w_{g}; X, Y_{g}) + \frac{\lambda_1}{2}\|w_g\|^2 + \mathcal{L}_{a}(w_{a}; X, Y_{a}) + \frac{\lambda_2}{2}\|w_a\|^2  + \frac{\lambda_3}{2}\mathcal{R}_{NOSSpaces}
\end{split}
    $}
\end{equation}
where $\mathcal{L}_{g}(w_{g}; X, Y_{g})$ refers to binary classification loss function in terms of gender classification, e.g., the widely used \emph{hinge loss function} \cite{steinwart2008support}, $\mathcal{L}_{a}(w_{a}; X, Y_{a})$ refers to the empirical loss of age regression, such as \emph{squares loss} \cite{hoerl1970ridge}, $\lambda_1$, $\lambda_2$, and $\lambda_3$ are trade-off parameters to balance the loss functions and regularizations. Obviously, by imposing the near orthogonality penalty on the semantic spaces, gender classification and age regression can be learned jointly with specially considering their semantic relationships.

\emph{It is worth noting that}
\begin{enumerate}
  \item Although the formulation \eqref{eq:R-NOSSpaces} is in form seemingly trivial, to the best of our knowledge, it is the first work in specifically exploring the semantic relationship between human gender and age.
  \item The NOSSpaces is similar \emph{in spirit} to but different \emph{in essence} from generic multi-task learning, because that in multi-task learning, different tasks are exclusively trained on their respective training data without data sharing, while in our framework the gender classification and age estimation are performed based on the same training data. More importantly, the formulation of NOSSpaces is relatively more concise than most of the multi-task learning methods.
  \item In theory, while there may be other strategies that can capture the semantic relationships between human gender and age, our joint learning methodology formulated in Eq. \eqref{eq:joint-framework} is concise and easy-to-implement, and more importantly, we later will experimentally demonstrate its effectiveness and superiority over related methods.
\end{enumerate}

\subsection{Joint Gender Classification (GC) and Age Estimation (AE) in the NOSSpaces} \label{sec:Joint GC&AE in NOSSpaces}
In order to evaluate the proposed framework and without loss of generality, we exemplify Eq. \eqref{eq:joint-framework} by taking the widely used binary-class SVM for gender classification, while the discriminant learning for ordinal regression (KDLOR) and support vector ordinal regression (SVOR) for age estimation, respectively.\\

\subsubsection{SVM for GC and Discriminant Learning for Ordinal Regression for AE in the NOSSpaces} \label{sec:SVM&RR&NOSSpaces}
With the proposed joint learning framework \eqref{eq:joint-framework}, we construct the first exemplified joint estimation model by respectively substituting the $\mathcal{L}_{g}(w_{g}; X_{g}, Y_{g})$ and the $\mathcal{L}_{a}(w_{a}; X_{a}, Y_{a})$ with the widely used binary SVM and the discriminant learning for ordinal regression (KDLOR) \footnote{Note that in order to follow the abbreviation, we still call its linear counterpart KDLOR, unless specified with linear or nonlinear description.} \cite{sun2010kernel} as
\begin{equation}\label{eq:SVM&KDLOR}
    \scalebox{\SCB}
    {$
\begin{split}
& min_{\{w_{g}, b_{g}, w_{a}, \rho\}} \\
& \quad \frac{1}{2}\|w_{g}\|^{2} + \lambda_{1}\sum_{i=1}^{N_{g}}max\{0, 1-y_{i}^{g}(w_{g}^{T}x^{g}_{i}+b_{g})\} \\
& \quad + w_{a}^{T}S_{w}w_{a} - \lambda_{2}\rho + \lambda_{3}(w_{g}^{T}w_{a})^{2} \\
& s.t. \\
& \quad w_{a}^{T}(m_{k+1} - m_{k}) \geq \rho, \; k = 1,2,...,K-1,
\end{split}
    $}
\end{equation}
where $N_{g}$ denotes the number of samples used for training in gender classification, $w_{g}$ and $b_{g}$, respectively, denote the weight vector and intercept for gender classification, $\{x_{i}^{g}, y_{i}^{g}\}_{i=1}^{N_{g}}$ denote the $i$-th instance and corresponding gender label of $N_{g}$ samples, $w_{a}$ denotes the weight vector for age estimation, $S_{w} \triangleq \frac{1}{N_{a}}\sum_{k=1}^{K}\sum_{x\in X_{k}}(x-m_{k})(x-m_{k})^{T}$ stands for the within-class scatter with $N_{a}$ being the total number of training samples of $K$ classes, $X_{k}$ the training samples set of the $k$-th class, and $m_{k}$ the mean vector of the $k$-th class, $\rho$ is interval margin between the classes, and $\lambda_{1}$, $\lambda_{2}$ and $\lambda_{3}$ are non-negative hyper-parameters.

Due to the bi-convexity of \eqref{eq:SVM&KDLOR} with respect to $\{w_{g}, b_{g}\}$ and $\{w_{a}\}$, i.e., it is convex with respect to $\{w_{g}, b_{g}\}$ with fixed $\{w_{a}\}$, and vice versa. Therefore, we can take an alternative strategy to solve $\{w_{g}, b_{g}, w_{a}\}$. More specifically, for fixed $w_{a}$, then \eqref{eq:SVM&KDLOR} becomes
\begin{equation}\label{eq:SVM&KDLOR-SVM}
    \scalebox{\SCB}
    {$
\begin{split}
& min_{\{w_{g}, b_{g}\}} \\
& \quad \frac{1}{2}\|w_{g}\|^{2} + \lambda_{1}\sum_{i=1}^{N_{g}}max\{0, 1-y_{i}^{g}(w_{g}^{T}x^{g}_{i}+b_{g})\} + \lambda_{3}(w_{g}^{T}w_{a})^{2}
\end{split}
    $}
\end{equation}
which is equivalent to
\begin{equation}\label{eq:SVM&KDLOR-SVM(equivalent)}
    \scalebox{\SCB}
    {$
\begin{split}
& min_{\{w_{g}, b_{g}\}} \\
& \quad \frac{1}{2}w_{g}^{T}(\mathcal{I}+2\lambda_{3}w_{a}w_{a}^{T})w_{g} + \lambda_{1}\sum_{i=1}^{N_{g}}max\{0, 1-y_{i}^{g}(w_{g}^{T}x^{g}_{i}+b_{g})\},
\end{split}
    $}
\end{equation}
where $\mathcal{I}$ is the identity matrix of proper size. The sub-problem \eqref{eq:SVM&KDLOR-SVM(equivalent)} is convex and can be similarly solved by the same way for SVM \cite{steinwart2008support}.

When the $w_{g}$ and $b_{g}$ are solved by \eqref{eq:SVM&KDLOR-SVM} or \eqref{eq:SVM&KDLOR-SVM(equivalent)}, then they are constant and we come to optimize \eqref{eq:SVM&KDLOR} with respect to $\{w_{a}\}$ as
\begin{equation}\label{eq:SVM&KDLOR-KDLOR}
    \scalebox{\SCB}
    {$
\begin{split}
& min_{\{w_{a}\}} \\
& \quad w_{a}^{T}S_{w}w_{a} - \lambda_{2}\rho + \lambda_{3}(w_{g}^{T}w_{a})^{2} \\
& s.t. \\
& \quad w_{a}^{T}(m_{k+1} - m_{k}) \geq \rho, \; k = 1,2,...,K-1,
\end{split}
    $}
\end{equation}
which in form is equivalent to
\begin{equation}\label{eq:SVM&KDLOR-KDLOR(equivalent)}
    \scalebox{\SCB}
    {$
\begin{split}
& min_{\{w_{a}\}} \\
& \quad w_{a}^{T}(S_{w}+\lambda_{3}w_{g}w_{g}^{T})w_{a} - \lambda_{2}\rho \\
& s.t. \\
& \quad w_{a}^{T}(m_{k+1} - m_{k}) \geq \rho, \; k = 1,2,...,K-1,
\end{split}
    $}
\end{equation}

Eq. \eqref{eq:SVM&KDLOR-KDLOR} in form is quite similar to the primal problem of KDLOR, and thus can be solved similarly as in \cite{sun2010kernel}.

By alternating between \eqref{eq:SVM&KDLOR-SVM(equivalent)} and \eqref{eq:SVM&KDLOR-KDLOR(equivalent)} until convergence, we can obtain the final solution for $w_{g}$, $b_{g}$, $w_{a}$, and $b_{a}$. And we summarize the complete procedure for joint SVM and KDLOR in Table \ref{tab: algorithm SVM-KDLOR}.
\begin{table}[!ht]
\centering
\caption{Algorithm of joint learning of SVM and KDLOR in the NOSSpaces.}\label{tab: algorithm SVM-KDLOR}
  \scalebox{1}
  {
\begin{tabular}{ll}
\hline
\hline
\MYnextline{l}{\textbf{Input:}\\ \\} & \MYnextline{l}{Training instances $X$, and labels $Y_{gender}$ and $Y_{age}$;\\ Parameters $\lambda_{1}$, $\lambda_{2}$, and $\lambda_{3}$.}\\
\textbf{Output:} & $w_{g}$, $b_{g}$, $w_{a}$.\\
\hline
\multicolumn{2}{l}{\MYnextline{l}{1.$\;$Initialize $w_{a}$;\\
2.$\;$\textbf{for} $t=1, 2, ...,T_{max}$ \textbf{do}\\
3.$\quad \ \ $ Compute $w_{g}$ and $b_{g}$ based on \eqref{eq:SVM&KDLOR-SVM(equivalent)};\\
4.$\quad \ \ $ Compute $w_{a}$ and $b_{a}$ based on \eqref{eq:SVM&KDLOR-KDLOR(equivalent)};\\
5.$\;$\textbf{end for}\\
6.$\;$Return $w_{g}$, $b_{g}$, and $w_{a}$.\\
}
} \\
\hline
\hline
\end{tabular}
   }
\end{table}

Using the $w_{g}$, $b_{g}$, and $w_{a}$ solved by the Algorithm in Table \ref{tab: algorithm SVM-KDLOR}, we can fulfil the goal of making joint estimation for human gender and age in the NOSSpaces.\\

\subsubsection{SVM for GC and Support Vector Ordinal Regression for AE in the NOSSpaces} \label{sec:SVM&SVOR&NOSSpaces}
In order to evaluate the general feasibility of the proposed joint learning framework. Besides the KDLOR, we also adopt the support vector ordinal regression (SVOR) \cite{chu2005new} method, a well-known ordinal regression method, for age estimation. And we can construct the corresponding joint model by substituting the formulations of SVM and SVOR into \eqref{eq:joint-framework} as
\begin{equation}\label{eq:SVM&SVOR}
    \scalebox{\SCB}
    {$
\begin{split}
& min_{\{w_{g}, b_{g}, w_{a}, b_{a} := \{b_{j}\}_{j=1}^{K}, \xi^{(*)}\}} \\
& \quad \frac{1}{2}\|w_{g}\|^{2} + \lambda_{1}\sum_{i=1}^{N_{g}}max\{0, 1-y_{i}^{g}(w_{g}^{T}x^{g}_{i}+b_{g})\} \\
& \quad + \frac{1}{2}\|w_{a}\|^{2} + \lambda_{2}\sum_{j=1}^{K}\sum_{i=1}^{N_{j}}(\xi_{i}^{j} + \xi_{i}^{j*}) + \lambda_{3}(w_{g}^{T}w_{a})^{2} \\
& s.t. \\
& \quad w_{a}^{T}x_{i}^{j} - b_{j} \leq -1 + \xi_{i}^{j}, \quad \xi_{i}^{j} \geq 0, \\
& \quad w_{a}^{T}x_{i}^{j*} - b_{j-1} \geq 1 - \xi_{i}^{j*}, \quad \xi_{i}^{j*} \geq 0,\\
& \quad b_{j-1} \leq b_{j},
\end{split}
    $}
\end{equation}
where $w_{g}$ and $b_{g}$, respectively, denote the weight vector and intercept for gender classification, $\{x_{i}^{g}, y_{i}^{g}\}_{i=1}^{Ng}$ denote the $i$-th instance and corresponding gender label of $N_{g}$ samples, $w_{a}$ and $b_{a}$, respectively, denote the weight vector and intercept for age estimation, $x_{i}^{j}$ stands for the $i$-th instance from the $j$-th age of totally $K$ ages, $\xi^{(*)}$ represent the slack variables, $b_{a} := \{b_{j}\}_{j=1}^{K}$ are the thresholds of SVOR to be optimized, and $\lambda_{1}$, $\lambda_{2}$ and $\lambda_{3}$ are also non-negative trade-off parameters.

Clearly, the formulation \eqref{eq:SVM&SVOR} is also bi-convex with respect to $\{w_{a}, b_{a}, \xi^{(*)}\}$ and $\{w_{g}, b_{g}\}$. Therefore, we also take the alternative optimization to solve it. More precisely, for fixed $\{w_{a}, b_{a}, \xi^{(*)}\}$, the \eqref{eq:SVM&SVOR} becomes
\begin{equation}\label{eq:SVM&SVOR-SVM}
    \scalebox{\SCB}
    {$
\begin{split}
& min_{\{w_{g}, b_{g}\}} \\
& \quad \frac{1}{2}\|w_{g}\|^{2} + \lambda_{1}\sum_{i=1}^{N_{g}}max\{0, 1-y_{i}^{g}(w_{g}^{T}x^{g}_{i}+b_{g})\} + \lambda_{3}(w_{g}^{T}w_{a})^{2}
\end{split}
    $}
\end{equation}
which is equivalent to
\begin{equation}\label{eq:SVM&SVOR-SVM(equivalent)}
    \scalebox{\SCB}
    {$
\begin{split}
& min_{\{w_{g}, b_{g}\}} \\
& \quad \frac{1}{2}w_{g}^{T}(\mathcal{I}+2\lambda_{3}w_{a}w_{a}^{T})w_{g} + \lambda_{1}\sum_{i=1}^{N_{g}}max\{0, 1-y_{i}^{g}(w_{g}^{T}x^{g}_{i}+b_{g})\},
\end{split}
    $}
\end{equation}
where $\mathcal{I}$ is the identity matrix of proper size. The sub-problem \eqref{eq:SVM&SVOR-SVM(equivalent)} is convex and can be similarly solved by the same way for SVM \cite{steinwart2008support}. Then, when the $\{w_{g}, b_{g}\}$ are obtained and fixed, the problem \eqref{eq:SVM&SVOR} becomes
\begin{equation}\label{eq:SVM&SVOR-SVOR}
    \scalebox{\SCB}
    {$
\begin{split}
& min_{\{w_{a}, b_{a}, \xi^{(*)}\}} \\
& \frac{1}{2}\|w_{a}\|^{2} + \lambda_{2}\sum_{j=1}^{K}\sum_{i=1}^{N_{j}}(\xi_{i}^{j} + \xi_{i}^{j*}) + \lambda_{3}(w_{g}^{T}w_{a})^{2} \\
& s.t. \\
& \quad w_{a}^{T}x_{i}^{j} - b_{j} \leq -1 + \xi_{i}^{j}, \quad \xi_{i}^{j} \geq 0, \\
& \quad w_{a}^{T}x_{i}^{j*} - b_{j-1} \geq 1 - \xi_{i}^{j*}, \quad \xi_{i}^{j*} \geq 0,\\
& \quad b_{j-1} \leq b_{j},
\end{split}
    $}
\end{equation}
or equivalently,
\begin{equation}\label{eq:SVM&SVOR-SVOR(equivalent)}
    \scalebox{\SCB}
    {$
\begin{split}
& min_{\{w_{a}, b_{a}, \xi^{(*)}\}} \\
& \frac{1}{2}w_{a}^{T}(\mathcal{I} + 2\lambda_{3}w_{g}w_{g}^{T})w_{a} + \lambda_{2}\sum_{j=1}^{K}\sum_{i=1}^{N_{j}}(\xi_{i}^{j} + \xi_{i}^{j*})\\
& s.t. \\
& \quad w_{a}^{T}x_{i}^{j} - b_{j} \leq -1 + \xi_{i}^{j}, \quad \xi_{i}^{j} \geq 0, \\
& \quad w_{a}^{T}x_{i}^{j*} - b_{j-1} \geq 1 - \xi_{i}^{j*}, \quad \xi_{i}^{j*} \geq 0,\\
& \quad b_{j-1} \leq b_{j},
\end{split}
    $}
\end{equation}
which is the same in form as the problem (5) of \cite{chu2005new}, and can be similarly solved as in the literature.

To obtain the final $\{w_{g}, b_{g}, w_{a}, b_{a}, \xi^{(*)}\}$, we repeat the alternative optimization process between \eqref{eq:SVM&SVOR-SVM(equivalent)} and \eqref{eq:SVM&SVOR-SVOR(equivalent)} until convergence, and summarize the algorithm in Table \ref{tab: algorithm SVM-SVOR}.
\begin{table}[!ht]
\centering
\caption{Algorithm of joint learning of SVM and SVOR in the NOSSpaces.}\label{tab: algorithm SVM-SVOR}
  \scalebox{1}
  {
\begin{tabular}{ll}
\hline
\hline
\MYnextline{l}{\textbf{Input:}\\ \\} & \MYnextline{l}{Training instances $X$, and labels $Y_{gender}$ and $Y_{age}$;\\ Parameters $\lambda_{1}$, $\lambda_{2}$, and $\lambda_{3}$.}\\
\textbf{Output:} & $w_{g}$, $b_{g}$, $w_{a}$, $b_{a}$.\\
\hline
\multicolumn{2}{l}{\MYnextline{l}{1.$\;$Initialize $w_{a}$, $b_{a}$, $\xi^{(*)}$;\\
2.$\;$\textbf{for} $t=1, 2, ...,T_{max}$ \textbf{do}\\
3.$\quad \ \ $ Compute $w_{g}$ and $b_{g}$ based on \eqref{eq:SVM&SVOR-SVM(equivalent)};\\
4.$\quad \ \ $ Compute $w_{a}$ and $b_{a}$ based on \eqref{eq:SVM&SVOR-SVOR(equivalent)};\\
5.$\;$\textbf{end for}\\
6.$\;$Return $w_{g}$, $b_{g}$, $w_{a}$, and $b_{a}$.\\
}
} \\
\hline
\hline
\end{tabular}
   }
\end{table}

When the $w_{g}$, $b_{g}$, $w_{a}$, and $b_{a}$ are obtained by the Algorithm listed in Table \ref{tab: algorithm SVM-SVOR}, we can make predictions for human gender and age.

\section{Joint Gender Classification and Age Estimation in the Nonlinear NOSSpaces} \label{sec:kernel-nosspaces}
\subsection{Nonlinear NOSSpaces} \label{sec:kernel-NOSSpaces}
In real applications, usually the associated data are not linearly separable. In order to handle such applications, we need to extend the joint learning framework in Eq. \eqref{eq:joint-framework} to higher-dimensional nonlinear spaces. Before that, we firstly give the specific representer theorem for the $w_g$ and $w_a$ involved in the joint learning framework as below.
\begin{thm}\label{Theorem:NOSSpaces-representation}
The $w_g$ and $w_a$ in Eq. \eqref{eq:joint-framework} can be respectively expressed as a linear combination of training samples as $w_g = \sum_{i}\alpha_{i}\phi(x^{i})$ and $w_a = \sum_{i}\beta_{i}\phi(x^{i})$, with $\alpha$ and $\beta$ being the combination coefficients and $\phi(\cdot)$ the feature mapping function defined on training samples $\{x_i\}^N_{i=1}$. (see the \hyperlink{proofREF}{Appendix} for the proof.)
\end{thm}
According to Theorem \ref{Theorem:NOSSpaces-representation}, the $w_g$ and $w_a$ can be respectively expressed as a combination of the training samples: $w_g = \sum_i\alpha_i\phi(x^i)$ and $w_g = \sum_i\beta_i\phi(x^i)$, and then NOSSpaces in Eq. \eqref{eq:R-NOSSpaces} can be mapped into the nonlinear feature space with kernel trick and expressed as
\begin{equation}\label{eq:kernel-NOSSpaces}
    \scalebox{\SCB}
    {$
\begin{split}
& \mathcal{R}_{Nonlinear-NOSSpaces}:= (\beta^T \textbf{K} \alpha)^{2} \\
& \; \; \; \quad \quad \quad \qquad \qquad \quad \quad = \alpha^T \textbf{K} \beta \beta^T \textbf{K} \alpha \\
& \; \; \; \quad \quad \quad \qquad \qquad \quad \quad = \beta^T \textbf{K} \alpha \alpha^T \textbf{K} \beta,
\end{split}
    $}
\end{equation}
where $\textbf{K}$ stands for the kernel matrix with $\textbf{K}_{(i,j)} = \phi(x^i)^T\phi(x^j)$.

\subsection{Joint GC and AE in the nonlinear NOSSpaces} \label{sec:GC&AE&kernel-NOSSpaces}
With Theorem \ref{Theorem:NOSSpaces-representation}, the joint learning models in Eq. \eqref{eq:SVM&KDLOR} can be reformulated in the nonlinear NOSSpaces as
\begin{equation}\label{eq:kernel:SVM&KDLOR}
    \scalebox{\SCB}
    {$
\begin{split}
& min_{\{\alpha, b_{g}, \beta, \rho\}} \\
& \quad \frac{1}{2}\alpha^T\textbf{K}\alpha + \lambda_{1}\sum_{i=1}^{N_{g}}max\{0, 1-y_{i}^{g}(\alpha^{T}\textbf{K}_{(:,i)}+b_{g})\} \\
& \quad + \frac{1}{N_a}\beta^T\textbf{KK}\beta - \lambda_{2}\rho + \lambda_{3}\alpha^T\textbf{K}\beta\beta^T\textbf{K}\alpha \\
& s.t. \\
& \quad \beta^T(\frac{1}{N_{k+1}}\textbf{K}_{(:,X_{k+1})}\textbf{1}_{k+1} - \frac{1}{N_{k}}\textbf{K}_{(:,X_{k})}\textbf{1}_{k}) \geq \rho, \; k = 1,2,...,K-1,
\end{split}
    $}
\end{equation}
with the assumption that the training samples have been centralized within each class beforehand, and $\textbf{K}_{(:,i)}$ represents the $i$-th column of the total kernel matrix $\textbf{K}$, and $\textbf{K}_{(:,X_{k})}$ stands for the sub-block kernel matrix between the $k$-th class samples and the entire training set, i.e., it is a sub-matrix of the $\textbf{K}$.

Similarly, the joint SVM and SVOR in Eq. \eqref{eq:SVM&SVOR} can be transformed into the form as
\begin{equation}\label{eq:kernel:SVM&SVOR}
    \scalebox{\SCB}
    {$
\begin{split}
& min_{\alpha, b_{g}, \beta, b_{a} := \{b_{j}\}_{j=1}^{K}, \xi^{(*)}\}} \\
& \quad \frac{1}{2}\alpha^T\textbf{K}\alpha + \lambda_{1}\sum_{i=1}^{N_{g}}max\{0, 1-y_{i}^{g}(\alpha^{T}\textbf{K}_{(:,i)}+b_{g})\} \\
& \quad + \frac{1}{2}\beta^T\textbf{K}\beta + \lambda_{2}\sum_{j=1}^{K}\sum_{i=1}^{N_{j}}(\xi_{i}^{j} + \xi_{i}^{j*}) + \lambda_{3}\beta^T\textbf{K}\alpha\alpha^T\textbf{K}\beta \\
& s.t. \\
& \quad \beta^T\textbf{K}_{(:,X_{i}^j)} - b_{j} \leq -1 + \xi_{i}^{j}, \quad \xi_{i}^{j} \geq 0, \\
& \quad  \beta^T\textbf{K}_{(:,X_{i}^{j*})} - b_{j-1} \geq 1 - \xi_{i}^{j*}, \quad \xi_{i}^{j*} \geq 0,\\
& \quad b_{j-1} \leq b_{j},
\end{split}
    $}
\end{equation}
where $\textbf{K}_{(:,X_{i}^{j(*)})}$ denotes the $i$-th column of the $\textbf{K}_{(:,X_{j})}$, where the meaning of $\textbf{K}_{(:,X_{j})}$ is the same as in Eq. \eqref{eq:kernel:SVM&KDLOR}.

By comparing the forms of Eqs \eqref{eq:kernel:SVM&KDLOR} with \eqref{eq:SVM&KDLOR}, \eqref{eq:kernel:SVM&SVOR} with \eqref{eq:SVM&SVOR}, respectively, it can be found that their forms are correspondingly similar, and thus the models in Eqs. \eqref{eq:kernel:SVM&KDLOR} and \eqref{eq:kernel:SVM&SVOR} can be solved similarly according to the Algorithms summarized in Tables \ref{tab: algorithm SVM-KDLOR} and \ref{tab: algorithm SVM-SVOR}, respectively.

\section{Experiment} \label{sec:experiment}
In this section, we conduct experiments to evaluate the proposed strategy on three benchmark aging datasets, respectively.
\subsection{Dataset} \label{sec:dataset}
\begin{figure}[htdp!]
  \centering
  \includegraphics[width=0.65\linewidth]{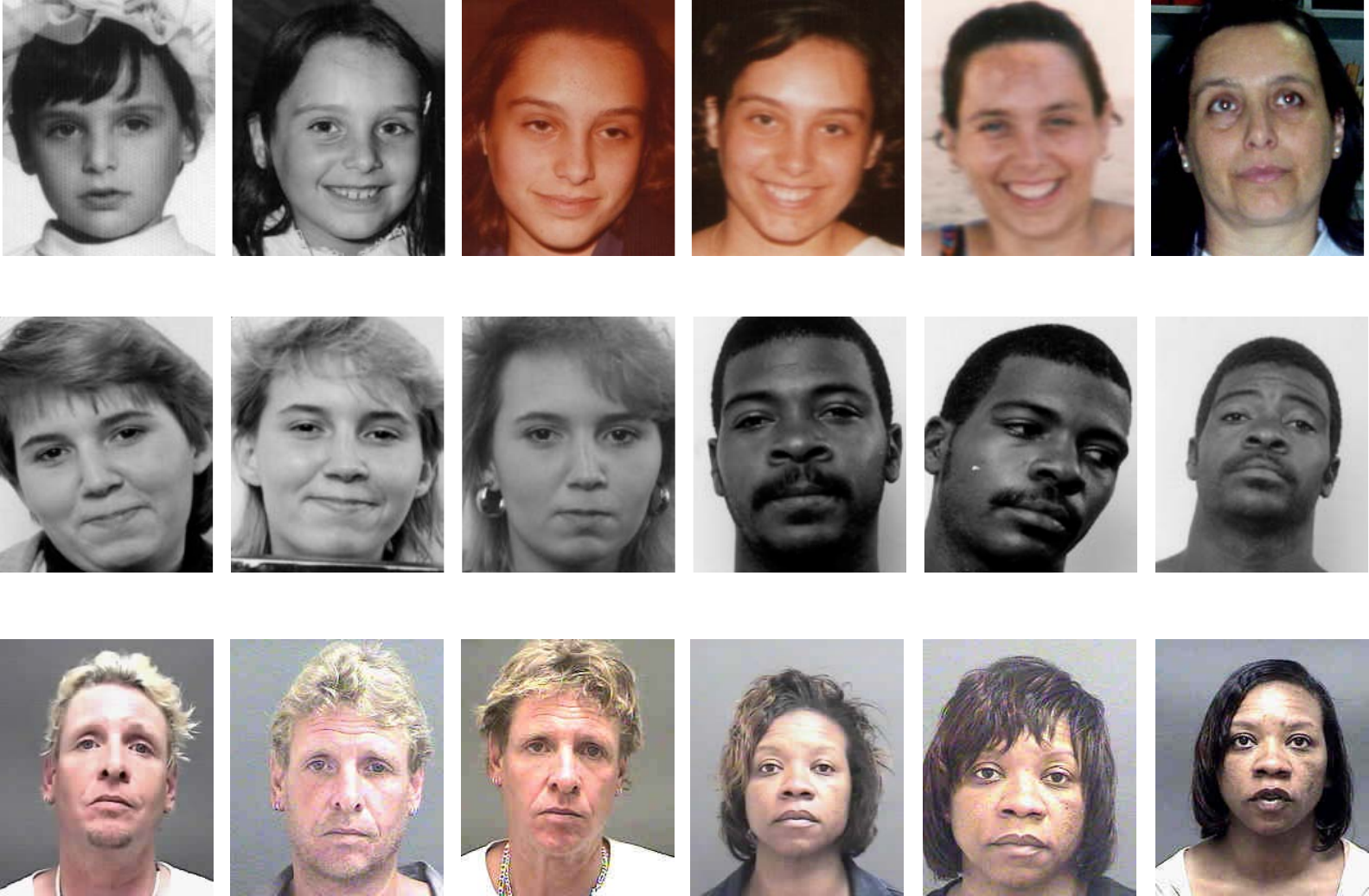}\\
  \caption{Image examples from the FG-NET (the first row), Morph Album I (the second row) and the Morph Album II (the third row).}\label{Fig:dataset-examples}
\end{figure}

In the experiments, we make evaluations of the proposed methods on three well-known aging datasets, FG-NET, Morph Album I, and Morph Album II. The FG-NET dataset consists of 1,002 facial images captured from 82 persons. In order to evaluate the proposed joint learning methods with respect to male and female age prediction, we select a subset of FG-NET with age ranging from 0 to 36 years old, since that there are very few female samples older than 36 years. For the Morph Album I dataset, there are about 1,690 facial images from about 631 persons aging from 16 to about 77 years old. And we select the subset from Morph Album I with age ranging from 16 to 44 years. Morph Album II is a relatively large aging dataset with over 55,000 images. We select the Caucasians from Album II with age ranging from 16 to 60 years for experiment. Image examples from the three datasets are shown in Figure \ref{Fig:dataset-examples}.

\subsection{Experimental Setup} \label{sec:setup}
With the three aging datasets, we extract AAM features for FG-NET and Morph Album I, BIF features for Morph Album II, respectively. Specifically, from the FG-NET and Morph Album I, we extract 200 dimensional AAM features from the FG-NET and Morph Album I, and 152 dimensional BIF features from the Album II for experiments, respectively.

In the experiments, the value of all the hyper-parameters involved in the work is set via \emph{cross-validation}. And it is worth noting that in order to make the semantic space of gender as orthogonal to that of age as possible, we assign the $\lambda_{3}$, in Eqs. \eqref{eq:SVM&KDLOR} and \eqref{eq:SVM&SVOR}, with a relatively large value, and in the experiments we tune it in \{$1e0$, $1e3$, $1e6$, $1e9$, $1e12$, $1e15$\}.

For performance measure, we uniformly adopt the commonly used classification \emph{Accuracy Rate} (Acc., $Acc.:= \frac{N_{correct}}{N_{total}}$ with $N_{correct}$ denoting the number of test samples correctly classified and $N_{total}$ the total number of test samples) for gender classification; Concerning age estimation, we take the \emph{Mean Absolute Error} (MAE, $MAE:= \frac{1}{N}\sum_{i=1}^{N}|\widehat{l}_i-l_i|$ with $l_i$ and $\widehat{l}_i$ denoting the ground-true and predicted age values, respectively) as the measure.

\subsection{Experimental Results and Analysis} \label{sec:experimental results}
\subsubsection{Linear Case} \label{sec:experimental results linear case}
\begin{figure*}[htp!]
  \centering
  \subfigure[FG-NET]{
    \label{fig:fgnet-gender} 
    \includegraphics[width=0.315\linewidth, height=1.2in]{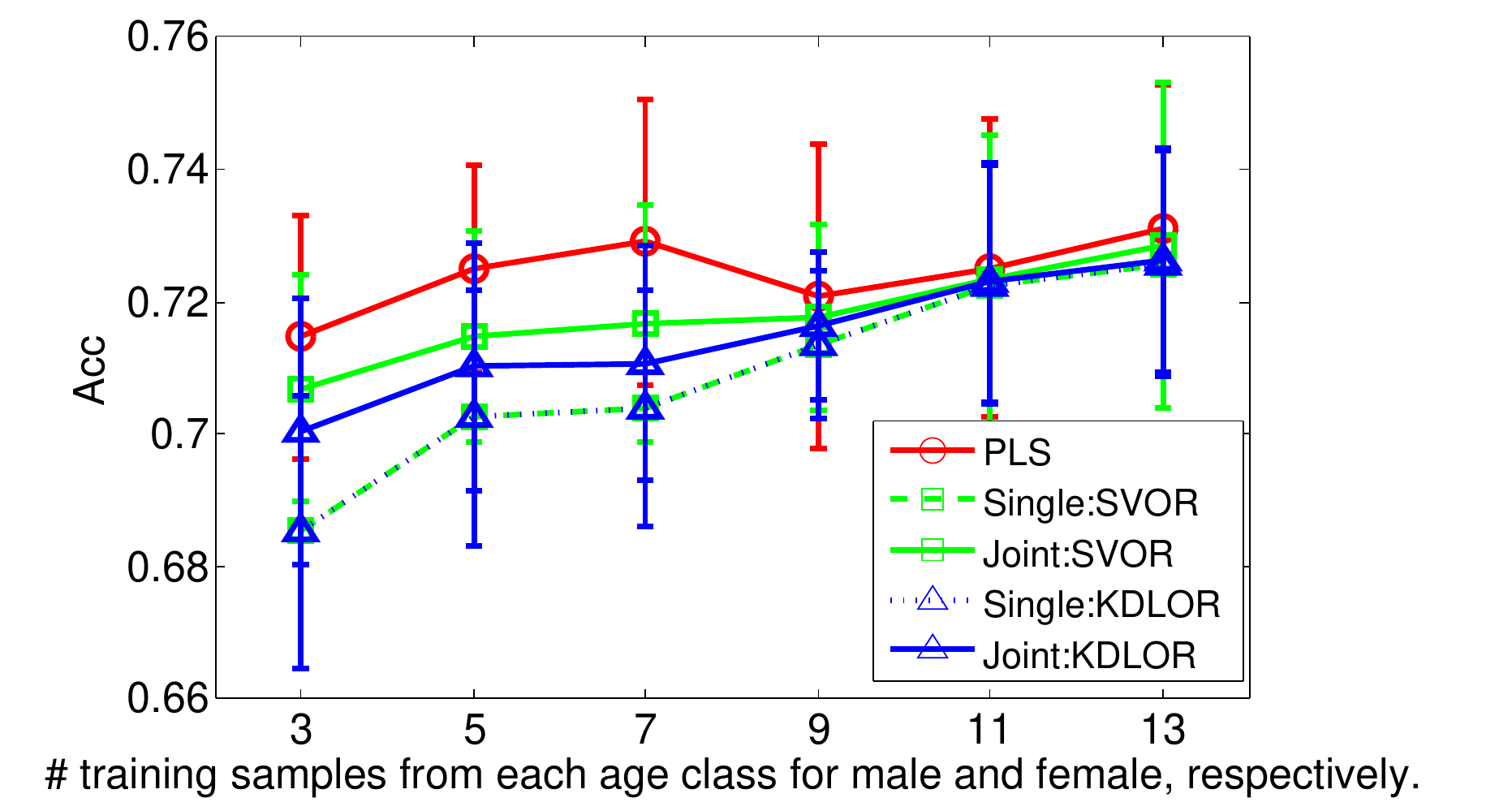}}
  \subfigure[Morph Album I]{
    \label{fig:alum1-gender} 
    \includegraphics[width=0.315\linewidth, height=1.2in]{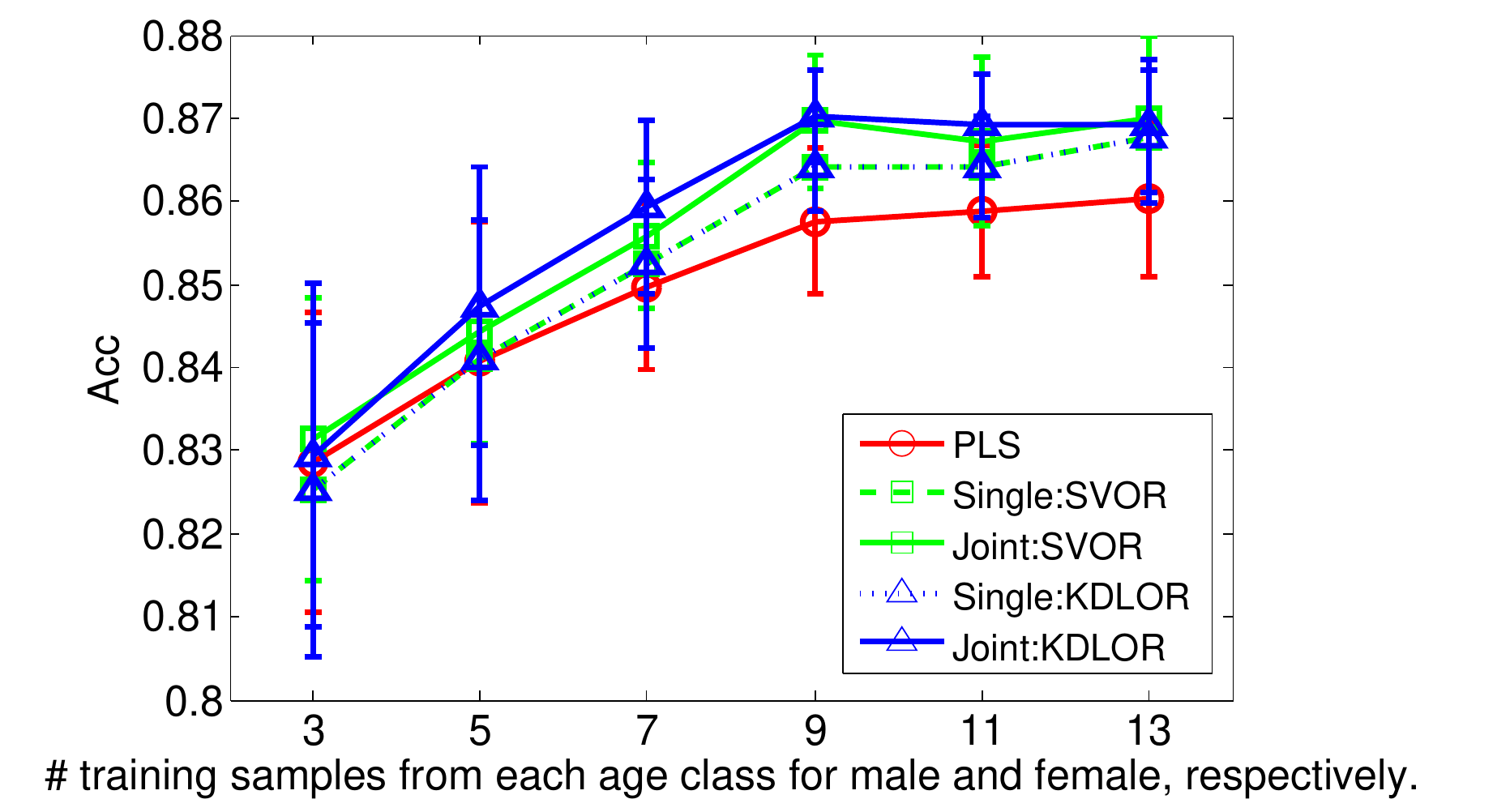}}
  \subfigure[Morph Album II]{
    \label{fig:alum1-gender} 
    \includegraphics[width=0.315\linewidth, height=1.2in]{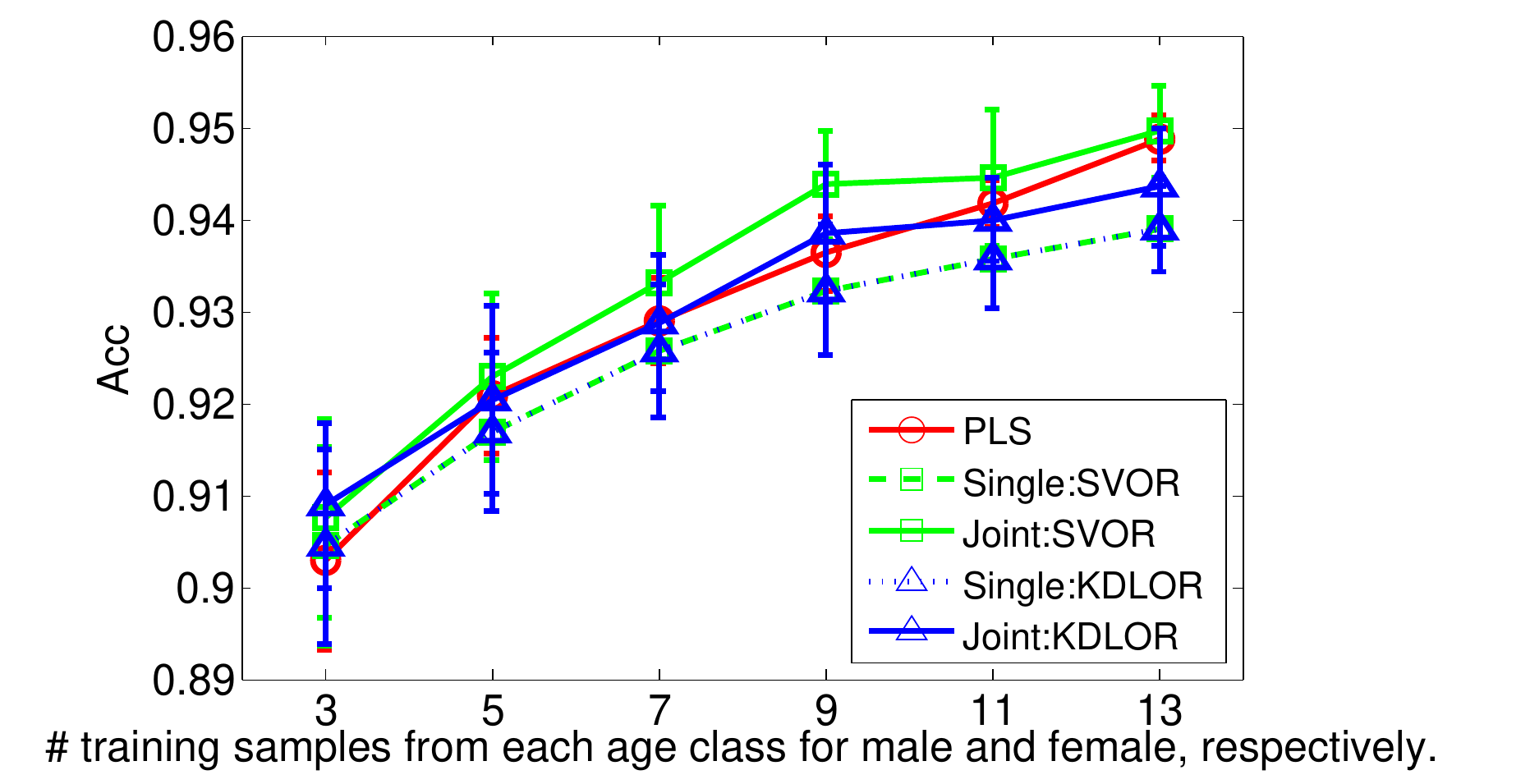}}
  \caption{Comparison between the methods in terms of gender classification in linear case.}
  \label{fig:gender-result} 
\end{figure*}

\begin{figure*}[htp!]
  \centering
  \subfigure[FG-NET]{
    \label{fig:fgnet-age} 
    \includegraphics[width=0.315\linewidth, height=1.2in]{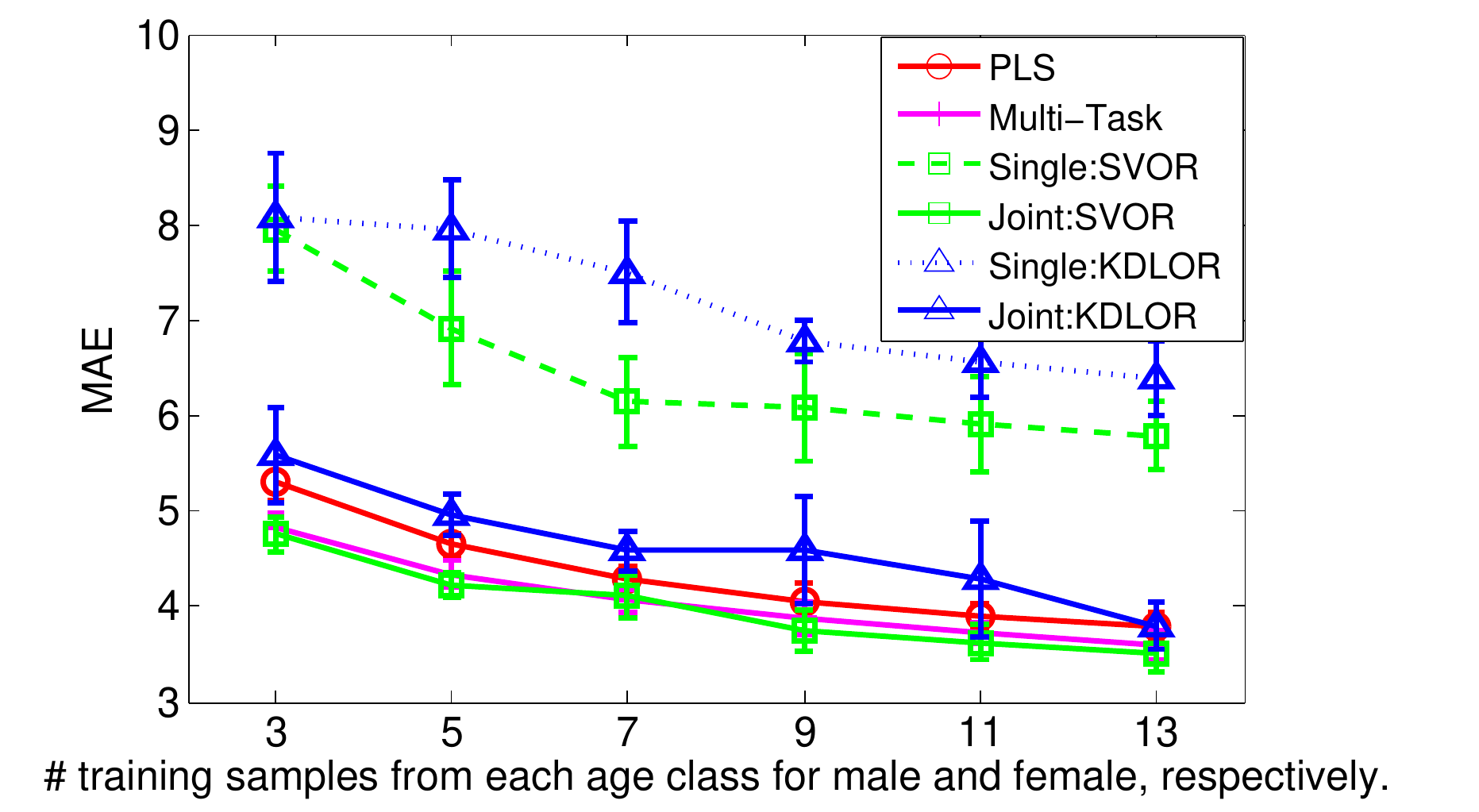}}
  \subfigure[Morph Album I]{
    \label{fig:alum1-age} 
    \includegraphics[width=0.315\linewidth, height=1.2in]{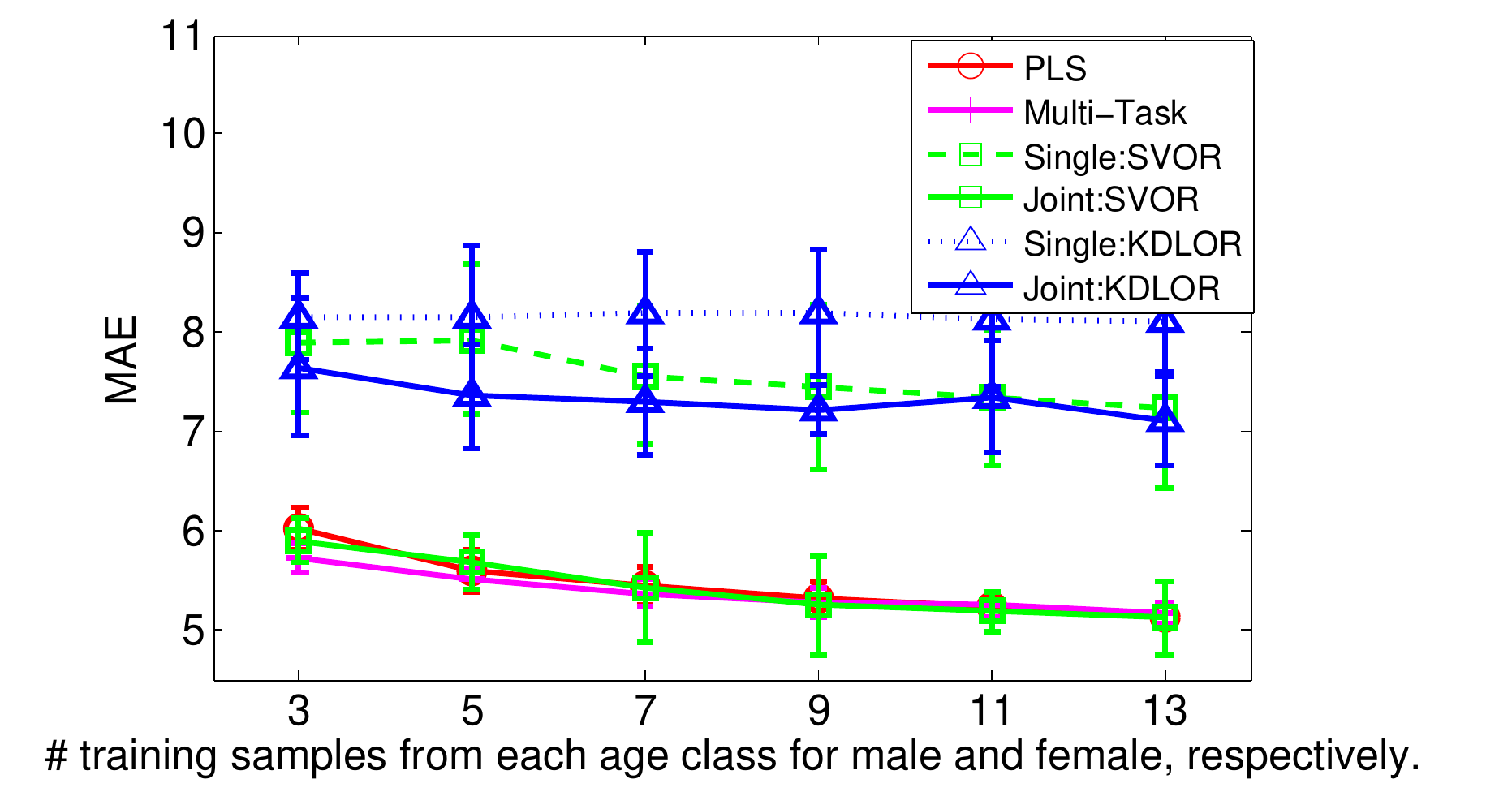}}
  \subfigure[Morph Album II]{
    \label{fig:alum1-age} 
    \includegraphics[width=0.315\linewidth, height=1.2in]{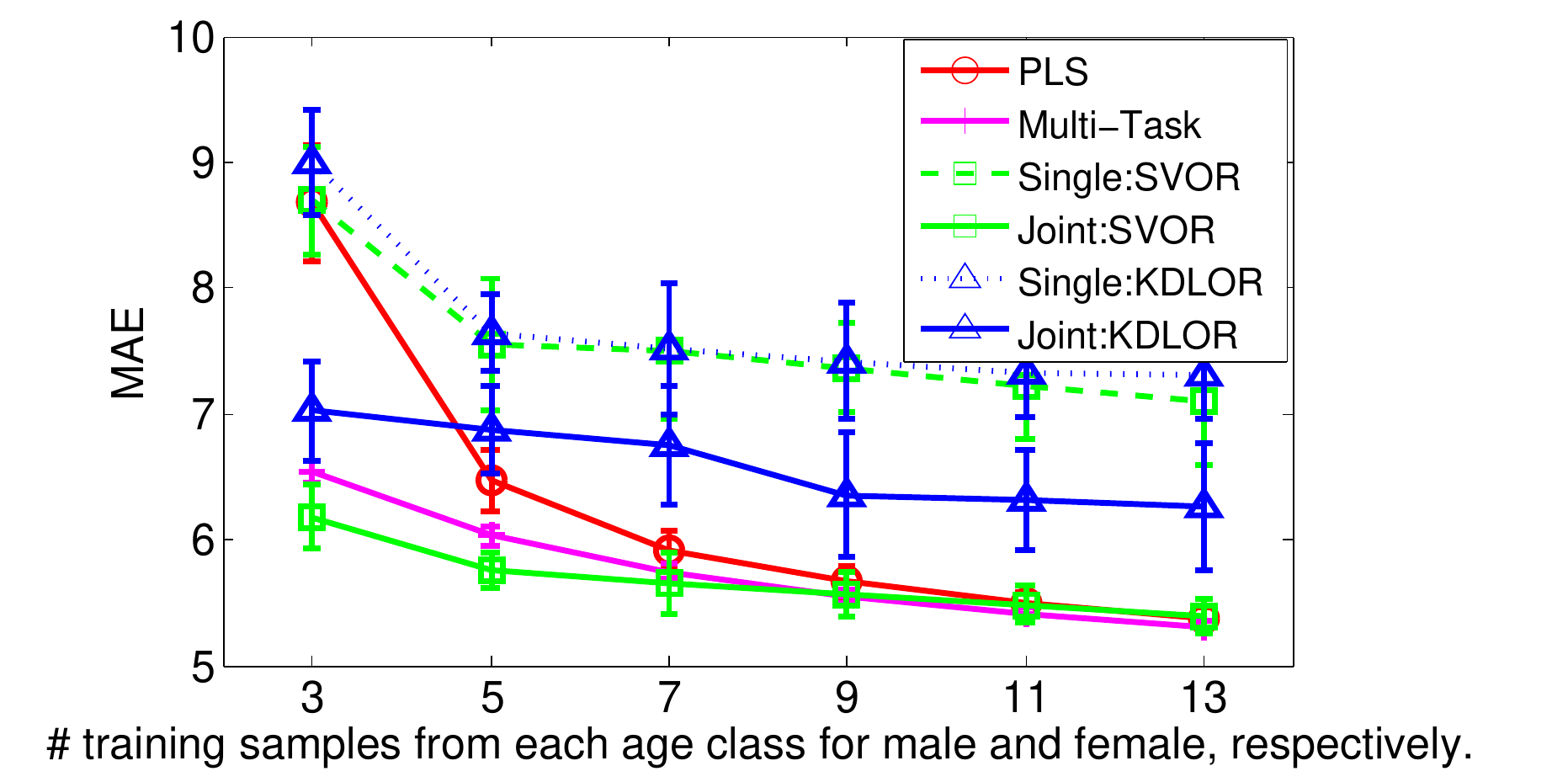}}
  \caption{Comparison between the methods in terms of age estimation in linear case.}
  \label{fig:age-result} 
\end{figure*}

With the extracted feature representations for FG-NET, Morph Album I and II, we randomly select a certain number of male and female samples from each age for training, with the remaining samples for test. And we demonstrate the experimental results in Figures \ref{fig:gender-result} and \ref{fig:age-result}.

From the results shown in Figure \ref{fig:gender-result}, it can be found that from the evaluation perspective of gender classification, joint learning based SVOR and KDLOR (in the NOSSpaces) can generate higher accuracy than their respectively single learning based counterparts. It demonstrates the effectiveness and superiority of learning in the NOSSpaces to gender recognition.

For age estimation, the learned regressors SVOR and KDLOR in the NOSSpaces can produce significantly lower MAEs than their jingle learning based methods, as shown in Figure \ref{fig:age-result}. More specifically, by the setting of joint learning in the NOSSpaces, the SVOR (KDLOR) can reduce the age estimation MAE by about 40\% (35\%), 30\% (10\%), and 25\% (over 10\%) on the FG-NET, Morph Album I and II, respectively. And in most cases, the joint learning based SVOR yields the lowest age estimation errors. These results demonstrate that learning in the NOSSpaces can dramatically improve the performance of human age estimation.

\subsubsection{Nonlinear Case} \label{sec:experimental results nonlinear case}
\begin{figure*}[htp!]
  \centering
  \subfigure[FG-NET]{
    \label{fig:fgnet-gender-kernel} 
    \includegraphics[width=0.315\linewidth, height=1.2in]{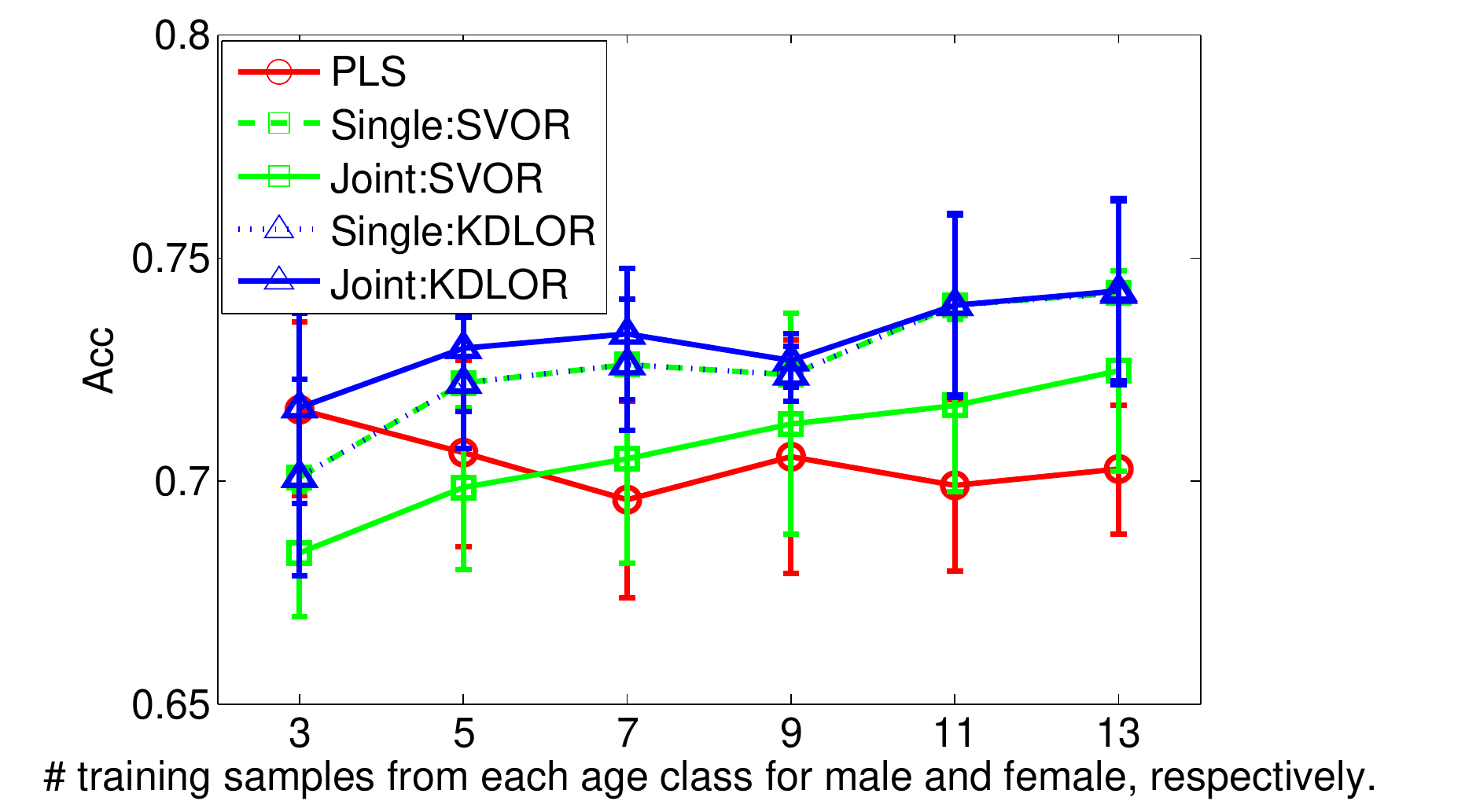}}
  \subfigure[Morph Album I]{
    \label{fig:alum1-gender-kernel} 
    \includegraphics[width=0.310\linewidth, height=1.2in]{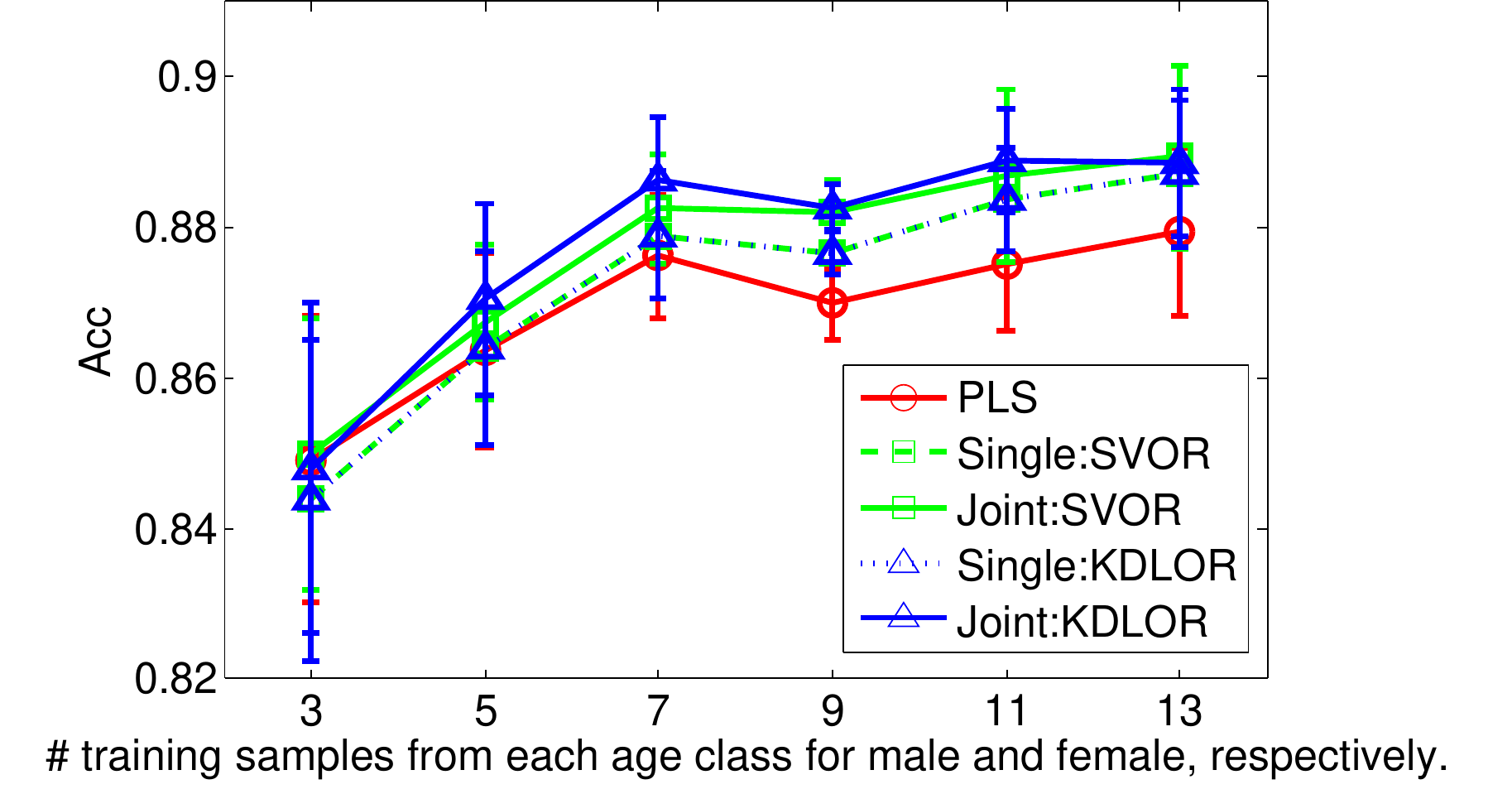}}
  \subfigure[Morph Album II]{
    \label{fig:alum1-gender-kernel} 
    \includegraphics[width=0.315\linewidth, height=1.2in]{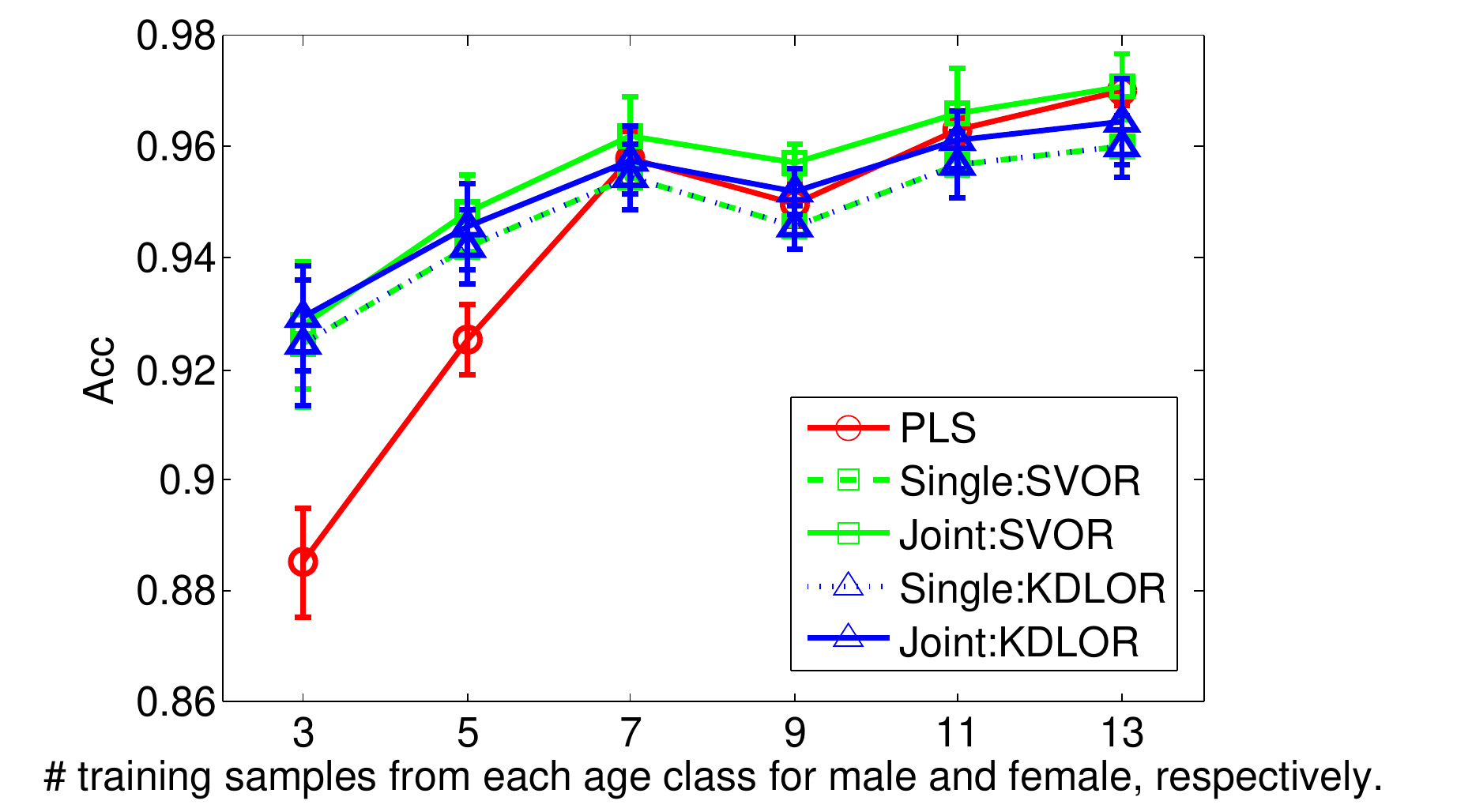}}
  \caption{Comparison between the methods in terms of gender classification in nonlinear case.}
  \label{fig:gender-result-kernel} 
\end{figure*}

\begin{figure*}[htp!]
  \centering
  \subfigure[FG-NET]{
    \label{fig:fgnet-age-kernel} 
    \includegraphics[width=0.315\linewidth, height=1.2in]{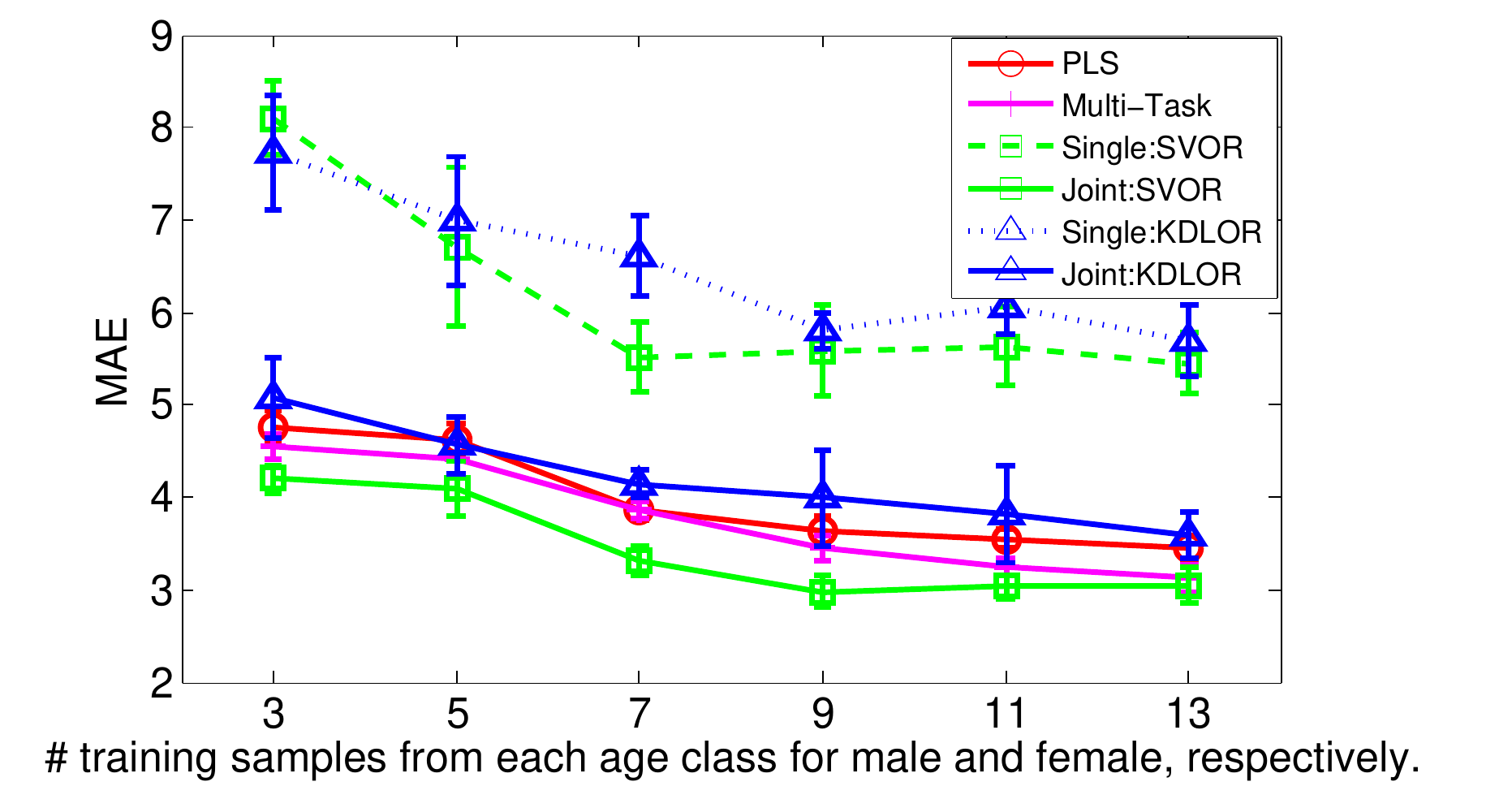}}
  \subfigure[Morph Album I]{
    \label{fig:alum1-age-kernel} 
    \includegraphics[width=0.315\linewidth, height=1.2in]{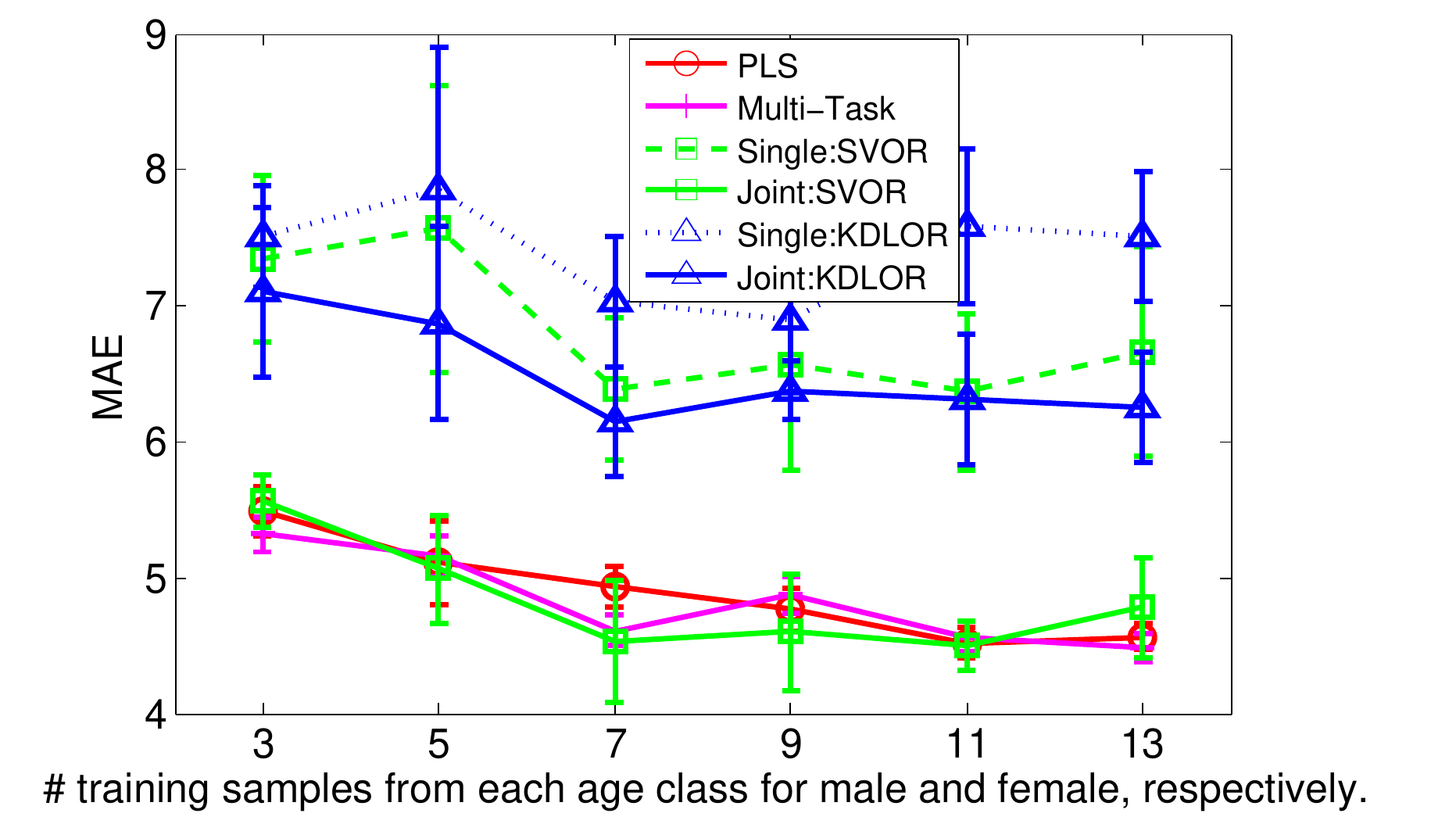}}
  \subfigure[Morph Album II]{
    \label{fig:alum1-age-kernel} 
    \includegraphics[width=0.315\linewidth, height=1.2in]{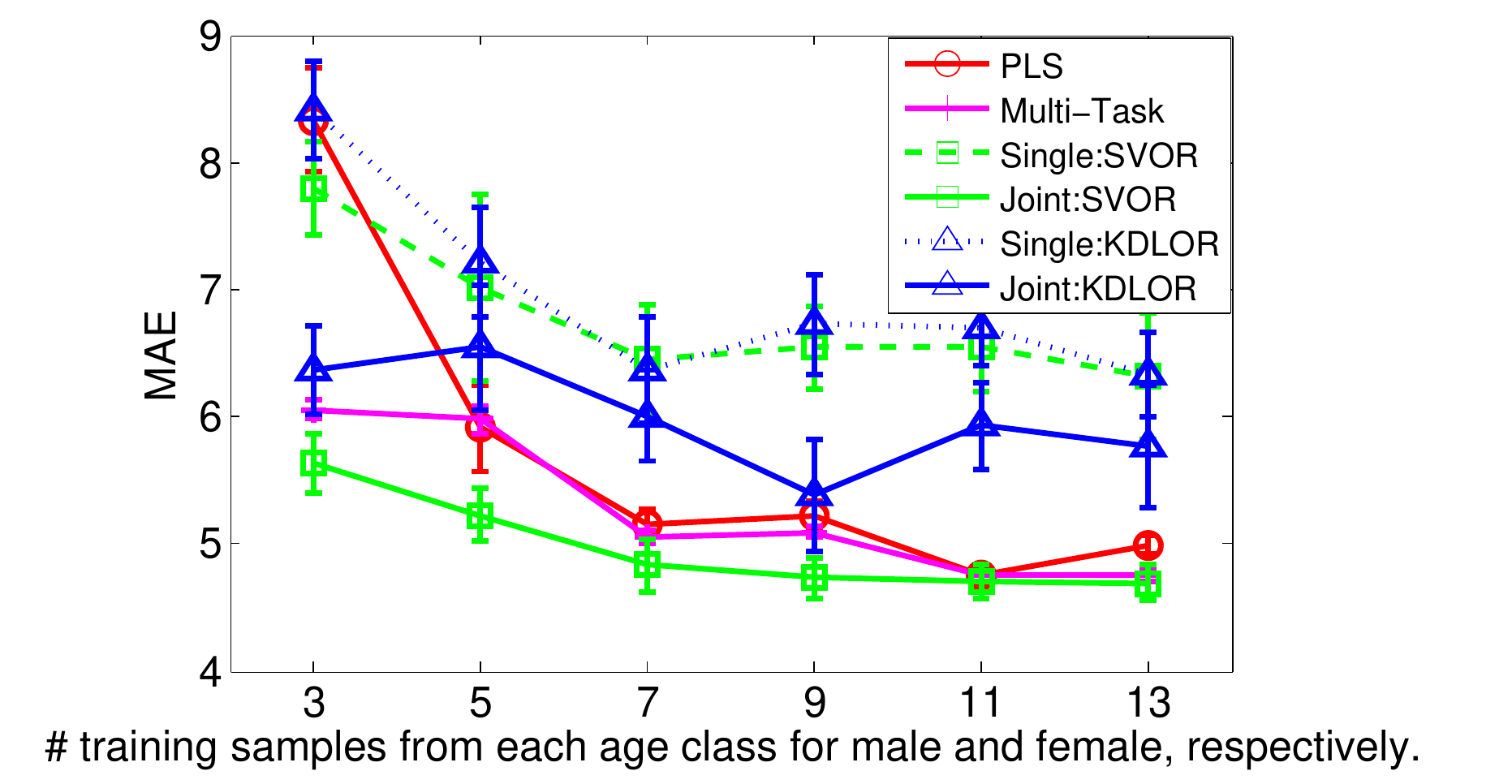}}
  \caption{Comparison between the methods in terms of age estimation in nonlinear case.}
  \label{fig:age-result-kernel} 
\end{figure*}

In order to evaluate the effectiveness of the proposed learning strategy in nonlinear feature space (i.e., the nonlinear NOSSpaces), we uniformly adopt the RBF kernel function to map the original feature representations of the three aging datasets (i.e., FG-NET, Morph Album I and Morph Album II) to high-dimensional feature space, and conduct experiments with results demonstrated in Figures \ref{fig:gender-result-kernel} and \ref{fig:age-result-kernel}.
%
By making comparison between Figures \ref{fig:gender-result-kernel} and \ref{fig:gender-result}, and between \ref{fig:age-result-kernel} and \ref{fig:age-result}, respectively in terms of gender classification and age estimation, we can find that the estimation accuracies obtained in the kernel-induced feature space are correspondingly higher than those in the original feature space. More specifically, for gender classification, the general average accuracy is increased by about 1\%, while the general age prediction error is reduced by over 6\% on average. And by making joint estimation in the kernel NOSSpaces, both the accuracies of gender classification and age estimation are further improved. It demonstrates the effectiveness of nonlinear NOSSpaces in improving the joint estimation of gender and age.

\subsection{Near-Orthogonality Between Human Gender and Age Semantic Spaces} \label{sec:angle analysis}
Here, in order to investigate the near-orthogonality between human gender and age semantic spaces, we denote the intersection angle between human gender semantic direction $w_g$ and age semantic direction $w_a$ as $\theta$
\begin{equation}\label{eq:angle between gender and age}
    \scalebox{\SCB}
    {$
\begin{split}
& \qquad cos(\theta) = \frac{<w_g, w_a>}{\|w_g\|\cdot\|w_a\|}. \\
\end{split}
    $}
\end{equation}
In the experiments, we find that generally, the $\theta$ lies near to $\frac{\pi}{2}$ \footnote{In the nonlinear NOSSpaces, the intersection angle between $\alpha$ and $\beta$ also lies near to $\frac{\pi}{2}$.}. That is, the gender direction $w_g$ is nearly but not strictly orthogonal to the age direction $w_a$. It witnesses the reasonableness of performing joint estimation for human gender and age in the nearly orthogonal semantic spaces (i.e., the NOSSpaces).

\subsection{Convergence Analysis} \label{sec:convergence analysis}
\begin{figure}[htbp!]
  \centering
  \includegraphics[width=0.68\textwidth]{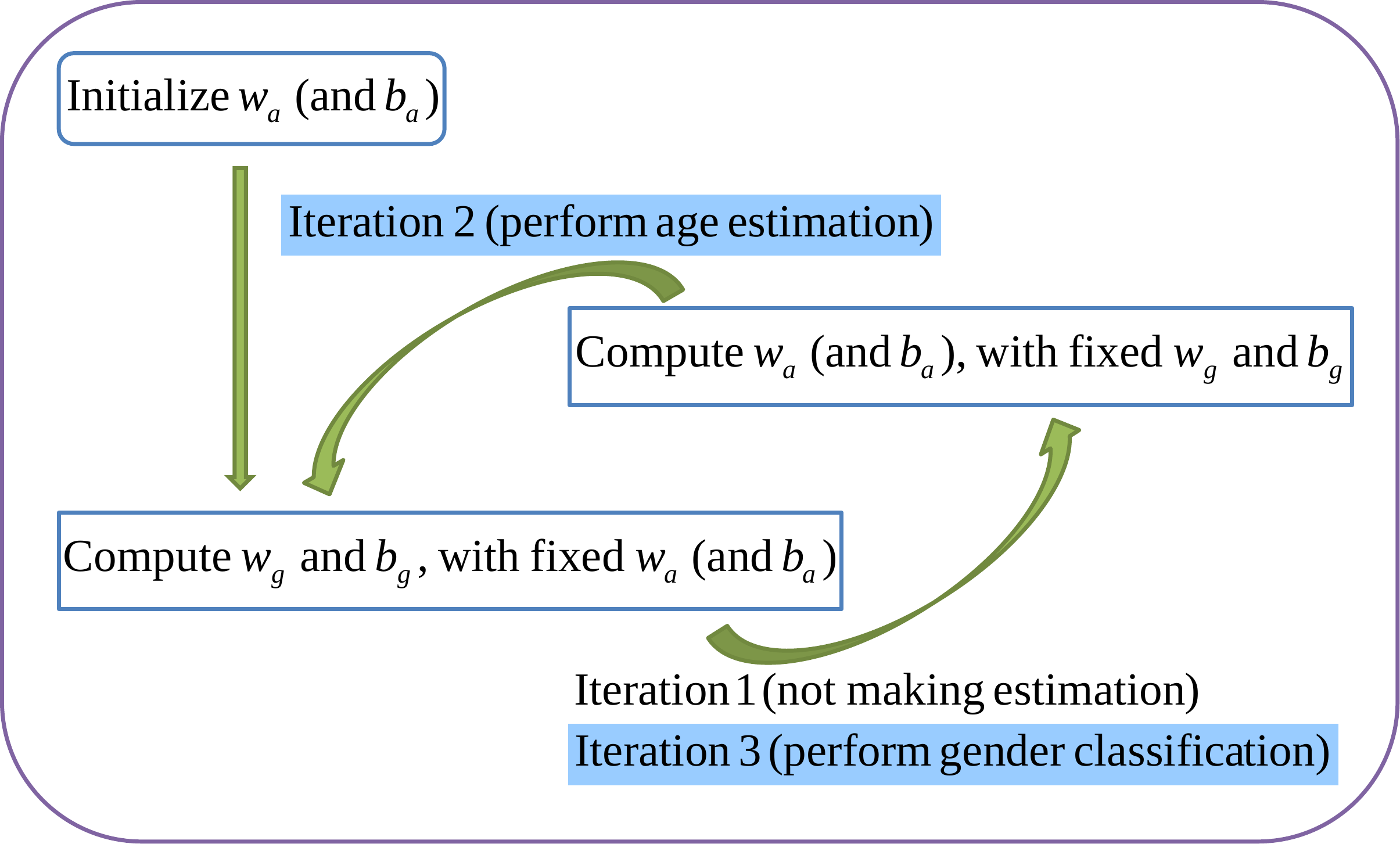}\\
  \caption{Convergence demonstration of the joint learning strategy for gender classification and age estimation.}\label{Fig:convergence-joint-learning-NOSSpaces}
\end{figure}

Through analyzing the algorithms for joint learning of SVM with KDLOR and SVM with SVOR, summarized in Tables \ref{tab: algorithm SVM-KDLOR} and \ref{tab: algorithm SVM-SVOR}, and performing the experiments, we find that it just requires 3 rounds of iterations for convergence of the joint human gender classification and age estimation in the NOSSpaces. For the sake of clarification, we provide an intuitive convergence analysis in Figure \ref{Fig:convergence-joint-learning-NOSSpaces}.

As shown in Figure \ref{Fig:convergence-joint-learning-NOSSpaces}, we make joint estimation for age estimation and gender classification in the second and third rounds of iterations of the joint learning strategy (in the NOSSpaces), respectively. And in the nonlinear NOSSpaces, it also performs age estimation and gender classification in the second and third round iterations, respectively. In other words, our nonlinear NOSSpaces just requires three rounds of iterations for the joint estimation task.

\section{Conclusions} \label{sec:conclusions}
In this work, we proposed a general framework for jointly estimating human gender and age, in which the binaryness of gender was considered by taking it as a binary classification problem, the ordinality of age was respected by treating it as an ordinal regression, and in particular, the semantic discrepancy between human gender and age was captured and expressed by nearly orthogonalizing their semantic spaces. In order to evaluate the proposed learning framework, we exemplified it by taking the widely used binary-class SVM for gender classification, while the discriminant learning for ordinal regression and support vector ordinal regression for age estimation, and then kernelized the joint learing framework by deriving a specific representer theorem. Finally, through experimental evaluations on three aging datasets, we demonstrated the effectiveness and superiority of the proposed methods. In the future, we consider to extend our methods to handle other joint estimation problems, such as joint estimation for human face-based age and expression, etc.

\section*{Acknowledgment} \label{sec:acknowledgment}
This work was partially supported by the National Natural Science Foundation of China under Grant $61472186$, the Specialized Research Fund for the Doctoral Program of Higher Education under Grant $20133218110032$, the Funding of Jiangsu Innovation Program for Graduate Education under Grant $CXLX13\_159$, and the Fundamental Research Funds for the Central Universities and Jiangsu \emph{Qing-Lan Project}.

\section*{Appendix} \label{sec:appendix-proof}
\hypertarget{proofREF}{}
\section*{The Proof of Theorem \ref{Theorem:NOSSpaces-representation}} \label{sec:proof}
\begin{proof}
Eq. \eqref{eq:joint-framework} can be detailed as
\begin{equation}\label{eq:kernel-joint-framework}
    \scalebox{\SCB}
    {$
\begin{split}
& \mathcal{J} = \mathcal{L}_{g}(w_{g}; X, Y_{g}) + \frac{\lambda_1}{2}\|w_g\|^2 + \mathcal{L}_{a}(w_{a}; X, Y_{a}) + \frac{\lambda_2}{2}\|w_a\|^2 + \frac{\lambda_3}{2}(w_{g}^{T}w_{a})^2.\\
& \;\;\; = \sum_i\mathcal{L}_{g}(w_{g}; x^i, y_{g}^i) + \frac{\lambda_1}{2}\|w_g\|^2 + \sum_i\mathcal{L}_{a}(w_{a}; x^i, y_{a}^i) + \frac{\lambda_2}{2}\|w_a\|^2 + \frac{\lambda_3}{2}(w_{g}^{T}w_{a})^2.\\
\end{split}
    $}
\end{equation}
Computing the derivatives of \eqref{eq:kernel-joint-framework} w.r.s. $w_g$ and $w_a$ and making them equal to zero leads to
\begin{subequations}  \label{eq:kernel-derivatives-WgWa}
\begin{align}
\frac{\partial \mathcal{J}}{\partial w_g} = \sum_{i}\mathcal{L}_{g}^{'}(w_{g}; x^i, y_{g}^i)x^i  + \lambda_1 w_g + \lambda_3(w_{g}^{T}w_{a})w_{a} = 0,   \label{eq:2A} \\
\frac{\partial \mathcal{J}}{\partial w_a} = \sum_{i}\mathcal{L}_{g}^{'}(w_{a}; x^i, y_{a}^i)x^i  + \lambda_2 w_a + \lambda_3(w_{g}^{T}w_{a})w_{g} = 0. \label{eq:2B}
\end{align}
\end{subequations}
Substituting \eqref{eq:2A} into \eqref{eq:2B} with $\lambda_2^{'} = \lambda_3(w_{g}^{T}w_{a})$ yields
\begin{equation}\label{eq:kernel-Wa1}
    \scalebox{\SCB}
    {$
\begin{split}
& \sum_{i}\mathcal{L}_{g}^{'}(w_{a}; x^i, y_{a}^i)x^i + \lambda_2^{'}(-\sum_{i}\mathcal{L}_{g}^{'}(w_{g}; x^i, y_{g}^i)x^i  - \lambda_1 w_g) + \lambda_2^{'}w_{g} = 0,\\
\end{split}
    $}
\end{equation}
which further yields
\begin{equation}\label{eq:kernel-Wa2}
    \scalebox{\SCB}
    {$
\begin{split}
& w_g = \frac{\lambda_2^{'}\sum_{i}\mathcal{L}_{g}^{'}(w_{g}; x^i, y_{g}^i)x^i - \sum_{i}\mathcal{L}_{g}^{'}(w_{a}; x^i, y_{a}^i)x^i}{(\lambda_2^{'} - \lambda_2^{'}\lambda_1)}.\\
\end{split}
    $}
\end{equation}
From Eq. \eqref{eq:kernel-Wa2}, it can be found that $w_g$ can be expressed by a linear combination of the training samples as
\begin{equation}\label{eq:kernel-Wa2}
    \scalebox{\SCB}
    {$
\begin{split}
& w_g = \sum_{i}\alpha_i x^i,\\
\end{split}
    $}
\end{equation}
, with
\begin{equation}\label{eq:kernel-Wa2}
    \scalebox{\SCB}
    {$
\begin{split}
& \alpha_i = \frac{\lambda_2^{'}\sum_{i}\mathcal{L}_{g}^{'}(w_{g}; x^i, y_{g}^i) - \sum_{i}\mathcal{L}_{g}^{'}(w_{a}; x^i, y_{a}^i)}{(\lambda_2^{'} - \lambda_2^{'}\lambda_1)}.\\
\end{split}
    $}
\end{equation}

Similarly, by substituting the $\lambda_1 w_g$ in \eqref{eq:2A} with \eqref{eq:2B}, we can similarly obtain a combination expression with the training samples.

Finally, introducing a feature mapping function $\phi(\cdot)$ on the samples $x^i$ along with the above proof can prove Theorem \ref{Theorem:NOSSpaces-representation}.

\end{proof}

\end{document}